\documentclass[letterpaper]{article} 
\usepackage{aaai2026}  
\usepackage{times}  
\usepackage{helvet}  
\usepackage{courier}  
\usepackage[hyphens]{url}  
\usepackage{graphicx} 
\usepackage{graphicx}
\usepackage{natbib}  
\usepackage{caption} 
\usepackage{algorithm}
\usepackage{algorithmic}
\usepackage{amsmath}
\usepackage{booktabs}
\usepackage{multirow}
\usepackage{pifont}
\usepackage{newfloat}
\usepackage{listings}
\DeclareCaptionStyle{ruled}{labelfont=normalfont,labelsep=colon,strut=off} 
\floatstyle{ruled}
\newfloat{listing}{tb}{lst}{}
\floatname{listing}{Listing}
\title{Beyond Plain Demos: A Demo-centric Anchoring Paradigm for In-Context Learning in Alzheimer's Disease Detection}
\author{
    Puzhen Su,
    Haoran Yin,
    Yongzhu Miao,
    Jintao Tang\textsuperscript{*},
    Shasha Li\textsuperscript{*},
    Ting Wang\thanks{Corresponding author.}
}
\affiliations{
    \textsuperscript{\rm 1}College of Computer Science and Technology, National University of Defense Technology, Changsha, China\\

    \{supuzhen, yinhaoran, miaoyz, tangjintao, shashali, tingwang\}@nudt.edu.cn
}

\usepackage{bibentry}

\begin{document}

\maketitle

\begin{abstract}
Detecting Alzheimer’s disease (AD) from narrative transcripts challenges large language models (LLMs): pre-training rarely covers this out-of-distribution task, and all transcript demos describe the same scene, producing highly homogeneous contexts. These factors cripple both the model’s built-in task knowledge (\textbf{task cognition}) and its ability to surface subtle, class-discriminative cues (\textbf{contextual perception}). Because cognition is fixed after pre-training, improving in-context learning (ICL) for AD detection hinges on enriching perception through better demonstration (demo) sets. We demonstrate that standard ICL quickly saturates, its demos lack diversity (context width) and fail to convey fine-grained signals (context depth), and that recent task vector (TV) approaches improve broad task adaptation by injecting TV into the LLMs' hidden states (HSs), they are ill-suited for AD detection due to the mismatch of injection granularity, strength and position. To address these bottlenecks, we introduce \textbf{DA4ICL}, a demo-centric anchoring framework that jointly expands context width via \emph{\textbf{Diverse and Contrastive Retrieval}} (DCR) and deepens each demo's signal via \emph{\textbf{Projected Vector Anchoring}} (PVA) at every Transformer layer. Across three AD benchmarks, DA4ICL achieves large, stable gains over both ICL and TV baselines, charting a new paradigm for fine-grained, OOD and low-resource LLM adaptation.
\end{abstract}

\begin{links}
    \link{Code}{https://github.com/Eneverg1veup/DA4ICL}
    \link{Datasets}{https://talkbank.org/dementia}
\end{links}

\section{Introduction}

Large language models (LLMs), trained on massive textual corpora, have exhibited impressive adaptability across diverse downstream tasks. Among various adaptation paradigms, in-context learning (ICL) has proven especially effective in low-resource settings~\cite{prompt_learning_1}, enabling LLMs to perform new tasks by conditioning on a handful of demonstrations (demos), without any parameter updates~\cite{icl_survey}. The effectiveness of ICL fundamentally depends on the \textit{task-related knowledge} acquired by LLMs during pre-training (\textbf{cognition}), and the \textit{text-to-label pattern} presented in the demos (\textbf{perception}), which together enable the model to infer task objectives and generalize from contextual cues~\cite{icl_mechanism_1}. In practice, the model’s task cognition is constrained by pre-training exposure~\cite{icl_mechanism_2}, making the construction (\textit{retrieval strategies}) and utilization (\textit{inference strategies}) of high-quality demo set critical for effective task perception. In this work, we view the core challenge of ICL as \textit{\textbf{how to optimally assemble limited demos to approximate the true task distribution, aggregating complementary task cues to maximize LLMs' perception}}.

Despite their remarkable adaptability, current ICL frameworks still encounter fundamental challenges in several practical scenarios, particularly in tasks characterized by low-resource conditions~\cite{OOD_2}, out-of-distribution (OOD) shifts~\cite{OOD_1}, and vague inter-class difference. A prime and societally significant example of such task is the early detection of Alzheimer’s disease (AD) from narrative speech or text~\cite{ad_bg2}. AD is a devastating neurodegenerative condition that affects millions worldwide with enormous personal and societal costs~\cite{ad_bg1,ad_bg3}. Critically, AD is currently incurable at advanced stages, making early and reliable detection essential for timely intervention and care. 
AD detection~\cite{ad_1} requires classifying the cognitive status (AD or healthy control, HC) of participants (PARs) based on their picture description transcripts. Yet, there are two parallel and intertwined properties that weaken both the cognition and perception of LLMs to achieve effective task adaptation. 
For \textbf{task cognition} (limited), due to privacy concerns and collection costs , AD detection has \textbf{extremely scarce and closed-source} datasets, making it a typical OOD task with minimal pre-training exposure. 
For \textbf{task perception} (poor), as all PARs describe the same scene, there exists \textbf{pervasive semantic homogeneity} caused by highly similar transcripts even across classes, contributing to weak and ambiguous text-to-label patterns in given demos.
Moreover, conventional ICL retrieval strategies are typically limited to a single aspect, most commonly semantic similarity~\cite{demo_sem_1,demo_sem_2}, resulting in demo sets that lack sufficient diversity and fail to provide the discriminative cues required for accurate AD detection. 
Similarly, existing inference strategies, like ensemble voting~\cite{ensemble_1,ensemble_2}, calibration~\cite{calibration_1}, are only effective when \textbf{\textit{task-related knowledge is sufficient or the demo set can optimally simulate the underlying task distribution}}—a condition rarely satisfied in AD scenario. As a result, robust task adaptation remains out of reach for current ICL methods, motivating the need for more expressive retrieval and demo enrichment mechanisms.

\begin{figure}[ht]
\centering
\includegraphics[height=6cm]{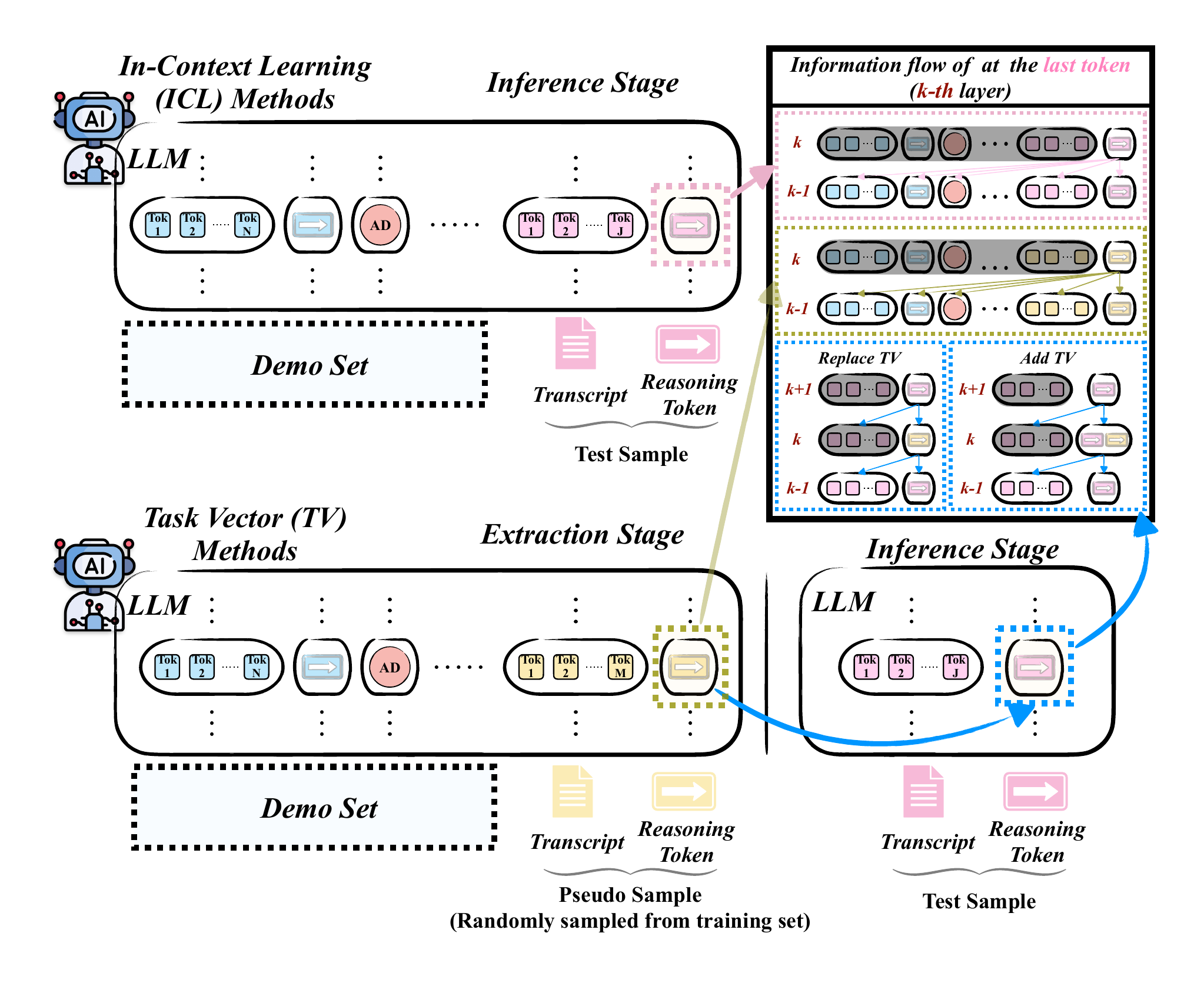} 
\caption{Schematic comparison of information flow and token processing in standard ICL versus TV methods.} 
\label{TV_mechanism}
\end{figure}

Recent works~\cite{tv_first} introduced Task Vector (TV) as a rapid task adaptation approach, which injects latent TV into the hidden states (HSs) of the test sample (at reasoning token, i.e., $\to$). TV methods typically split ICL into two stages: \textbf{1) extraction}, where a demo set is concatenated with a randomly selected pseudo sample and the last $\to$ token’s HS is extracted as the TV, and \textbf{2) inference}, where the label for a test sample is predicted by injecting this vector into its $\to$ token. While effective for generic tasks\cite{tv_2}, TV methods present two fundamental limitations for fine-grained tasks such as AD detection. First, in ICL, each demo serves as an anchor~\cite{label_words,anchor}, aggregating semantic information and guiding the final prediction. However, existing TV paradigms are fundamentally \textbf{test-sample-centric} (see Fig.~\ref{TV_mechanism}), they inject adaptation signals at the final $\to$ token (i.e., in test sample), discarding the distributed information encoded in preceding demo anchors and instead relying on a single, pseudo-sample-derived TV. Such injection paradigm neglects context diversity and fine-grained cues which are crucial in tasks with subtle inter-class variation. Second, most TV methods inject TVs via \textbf{addition or replacement}, directly altering both the \textbf{\textit{direction}} and \textbf{\textit{magnitude}} of the HS, which may risk distorting the original semantic information and introduce instability or bias. Consequently, \textbf{\textit{existing TV approaches often fail to deliver reliable performance on OOD, low-resource, and fine-grained tasks like AD detection}}.

\begin{figure}[!ht]
\centering
\includegraphics[height=5.5cm]{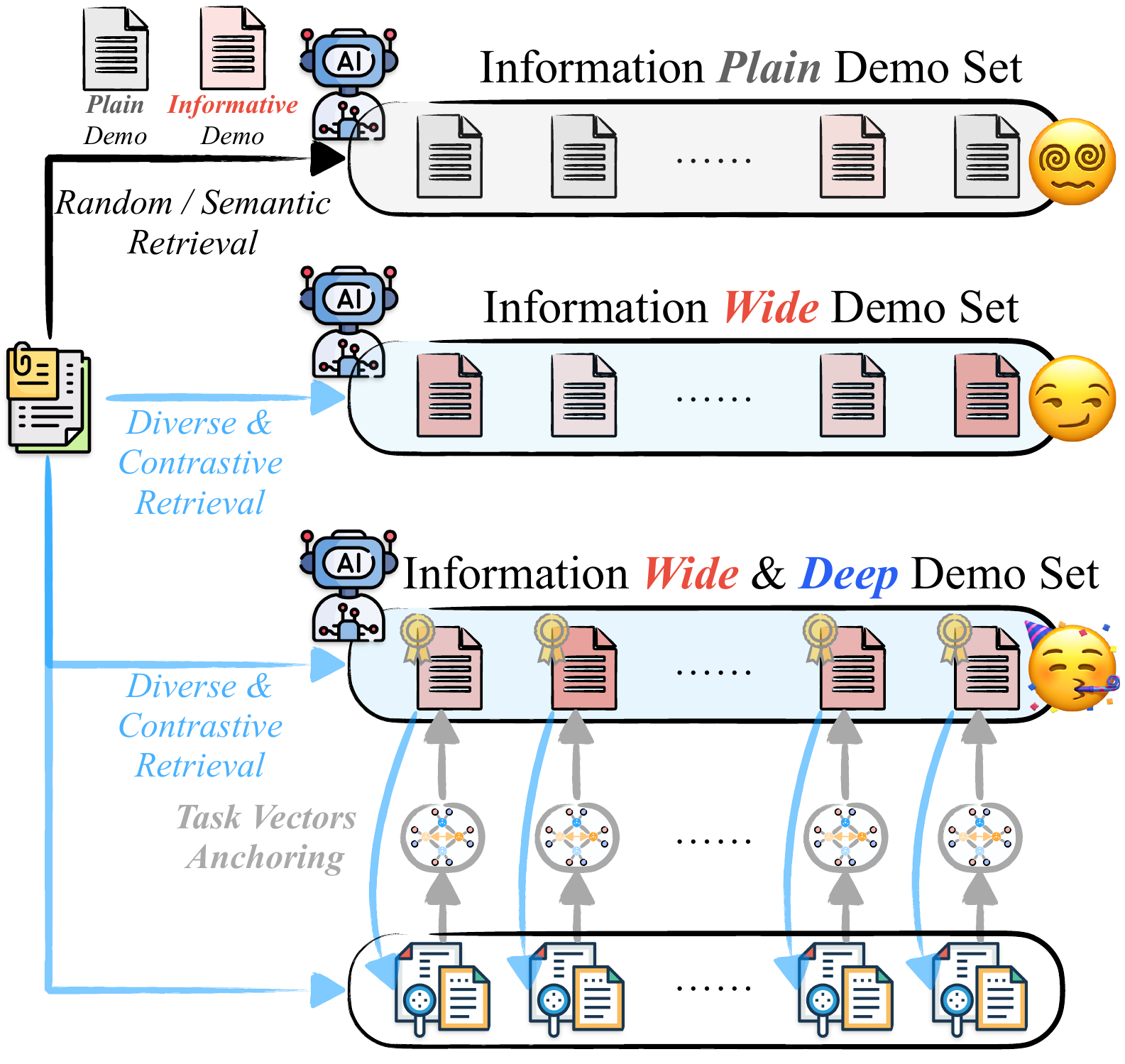} 
\caption{Progressive enrichment of demo sets, from plain to wide and deep, drives more effective and robust in-context reasoning.}
\label{fig_first}
\end{figure}

To overcome these limitations of both ICL and TV paradigms, we propose \textit{Demo-centric Anchoring for In-Context Learning} (DA4ICL), a paradigm shift that rethinks how demo sets are constructed and how task information is integrated. Our approach is driven by two core motivations: (1) \textbf{\textit{maximizing demo context width by diversifying and contrasting the information available to LLM}}, and (2) \textbf{\textit{enhancing context depth by reinforcing each demo with fine-grained and demo-centric signals}}.
As shown in Fig.~\ref{fig_first}, DA4ICL first employs a \textit{Diverse and Contrastive Retrieval} (DCR) strategy to construct a context-wide \textit{main-demo} set, selecting demos from two complementary perspectives (i.e., semantic, structural), and broadening the contextual cues available for adaptation. Next, for each \textit{main-demo}, a second-stage retrieval process identifies a set of \textit{sub-demos}, enabling us to extract detailed, demo-specific TVs. We then introduce \textit{Projected Vector Anchoring} (PVA) mechanism, which projects these fine-grained TVs into corresponding $\to$ tokens of each \textit{main-demo} across all Transformer layers, treating each demo as an anchor for subsequent reasoning. Unlike previous test-centric TV approaches, DA4ICL leaves the test sample token unmodified. During inference, the LLM naturally aggregates information from all enriched demo anchors through masked self-attention, thereby enhancing both stability and precision. By shifting from a single, test-centric injection to a distributed, demo-centric anchoring paradigm, DA4ICL substantially increases the diversity and discriminative value of context available to the LLM. This enables LLM to leverage more nuanced cues and improves task perception by addressing both context width and depth bottlenecks. Experimental results across three AD detection datasets confirm that our method consistently and significantly outperforms both conventional ICL and TV baselines, establishing a new paradigm for effective adaptation in challenging NLP tasks. Our main contributions are:
\begin{itemize}
\item We propose a demo-centric anchoring paradigm that shifts task information integration from the test token to each demo anchor, enhancing fine-grained in-context reasoning.
\item We introduce a diverse and contrastive retrieval strategy (DCR) to maximize context width, capturing multi-dimensional and complementary contextual cues for robust adaptation.
\item We design a projection-based, layer-wise anchoring mechanism (PVA) that deepens context integration by injecting fine-grained task vectors into demo anchors without distorting original semantics.
\end{itemize}

\begin{figure*}[t]
\centering
\small
\includegraphics[width=0.9\textwidth]{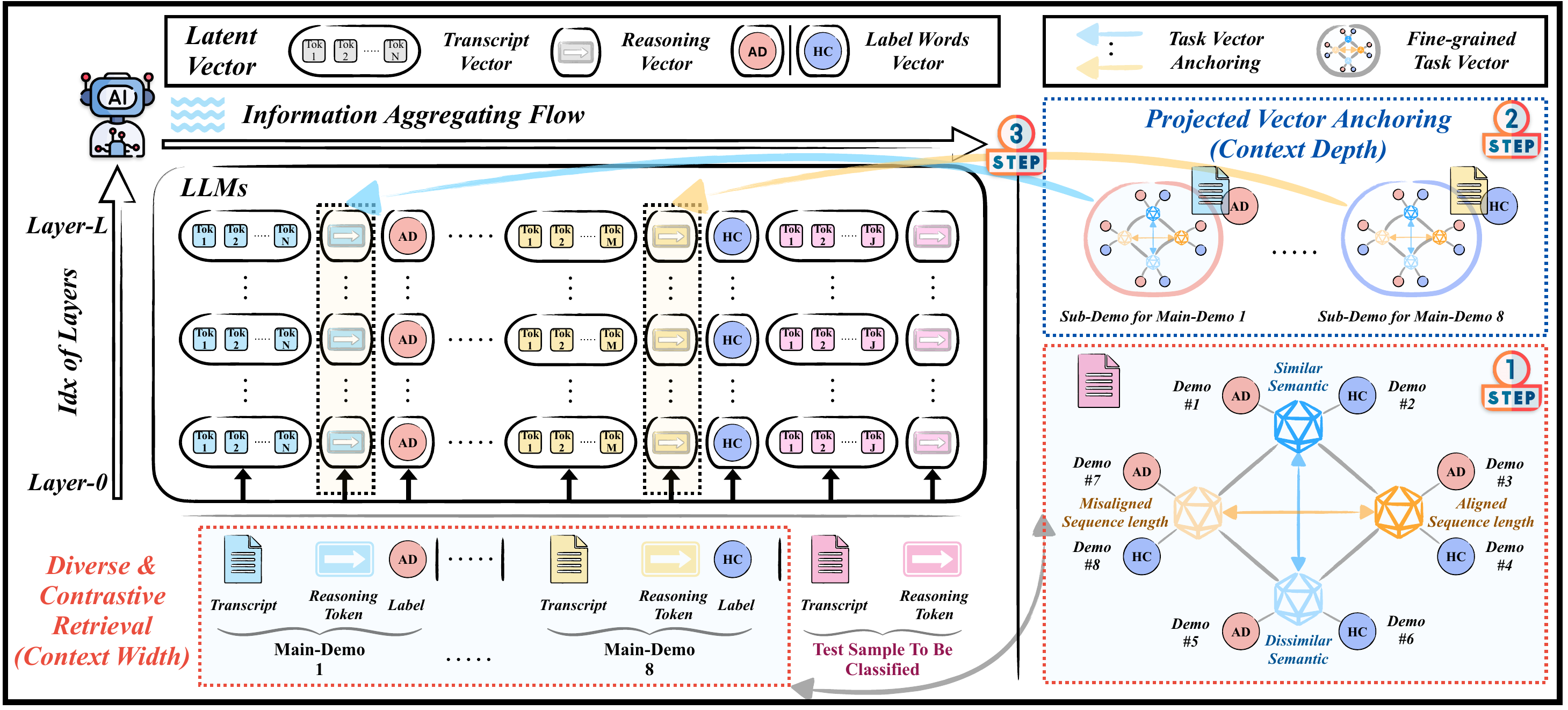} 
\caption{Overview of the DA4ICL framework. Diverse and contrastive demos are selected and enriched via projected vector anchoring across all Transformer layers to provide both wide and deep context for robust AD detection.}
\label{fig1}
\end{figure*}

\section{Related Works}

\paragraph{In-Context Learning Methods.}
Recent studies~\cite{icl_mechanism_1, icl_mechanism_2, icl_mechanism_3} have highlighted two aspects determining ICL effectiveness in LLMs: the recognition of task objectives (\textbf{task cognition}) and leveraging relevant contextual cues (\textbf{task perception}). Task cognition, defined as the latent task-specific knowledge acquired during pre-training, fundamentally constrains ICL adaptation, when tasks lack sufficient pre-training exposure, models often default to superficial label copying from semantically similar demos~\cite{icl_copy}. Furthermore, in view of information flow, Transformer models heavily rely on aggregating semantic signals at the final token’s HS for inference~\cite{label_words,anchor}, making the diversity and informativeness of demos critical for effective adaptation. To address these limitations, previous works primarily explored retrieval strategies based on semantic similarity~\cite{demo_sem_1,demo_sem_2} and ensemble-based inference~\cite{ensemble_3, ensemble_4}. However, these approaches remain fundamentally constrained in OOD and fine-grained tasks like AD detection, where semantically homogeneous inputs and subtle class distinctions exacerbate the bottlenecks of existing ICL paradigms on AD detection~\cite{icl_ad_1,icl_ad_2}.

\paragraph{Task Vector Methods.}
TV methods~\cite{tv_1} have recently emerged as an alternative to ICL, by injecting a task-specific vector directly into LLM’s HSs. In these approaches~\cite{tv_2}, an LLM’s few-shot demos are first compressed into a single TV, typically by extracting the HS at the reasoning token (i.e. the final separator token $\to$ in a prompt that maps inputs to outputs). This TV is then injected at the corresponding position during test sample’s forward pass. While effective for broad tasks~\cite{tv_sv, tv_fv}, TV methods face fundamental limitations in low-resource, OOD, and fine-grained settings like AD detection. First, existing TV paradigms are based on single-layer, last-token injection~\cite{tv_r_1}, assuming the extracted TV captures the full task context, which is rarely satisfied for tasks that require subtle or distributed cues. Pseudo-sample overfitting~\cite{tv_first} is another key issue, where TV injection can overfit to the idiosyncrasies of the pseudo-sample, neglecting important distinctions. This is particularly problematic for AD detection, where nuanced inter-class variations are crucial. Second, TV injection typically involves addition~\cite{tv_icv} or replacement~\cite{separator_token_1}, modifying both the \textbf{\textit{direction}} and \textbf{\textit{magnitude}} of the HS. This often leads to semantic misalignment, distorting the original meaning and introducing instability or bias, especially when the injected TV is poorly aligned with the context. These challenges underscore the need for fine-grained, demo-aware, and projecting-based injection strategies.

\section{Methodology}
\paragraph{Preliminary: Why Test-centric TV Injection Fails.}
In decoder-only LLMs, each token's HS is updated layer by layer via residual connections~\cite{residual_1,residual_2}:
\begin{equation}
\small
h_t^{(\ell)} = h_t^{(\ell-1)} + \mathrm{mha}(h_t^{(\ell-1)}) + \mathrm{mlp}(h_t^{(\ell-1)} + \mathrm{mha}(h_t^{(\ell-1)})),
\end{equation}
where $\mathrm{mha}$ and $\mathrm{mlp}$ denote multi-head attention and feed-forward modules. This recursive structure enables each layer to integrate contextual signals from preceding tokens and layers. At inference, the LLM predicts the next token $x_{t+1}$ via:
\begin{equation}
\small
P(x_{t+1} \mid x_{\le t}) = \mathrm{softmax}(\mathbf{W}_{\mathrm{LM}} \cdot h_t^{(L)}),
\end{equation}
where $\mathbf{W}_{\mathrm{LM}}$ is the unembedding matrix that maps the last token’s final-layer HS $h_t^{(L)}$ to vocabulary logits. This encourages existing TV methods to inject TVs into the test $\to$ token at a single layer. However, this \textit{test-centric} injection is coarse-grained and prone to overfitting in fine-grained tasks like AD detection. Since the $\to$ tokens in demos act as local anchors encoding class-specific cues, injecting signals into them can further enhance demo representation, but shallow-layer and signals often dissipate through the residual stream. To preserve their influence, TVs must be anchored across all layers. These insights motivate our \textbf{demo-centric, full-layer anchoring} design.

\paragraph{DA4ICL Framework.}
To address the intrinsic information bottleneck and granularity misalignment inherent in standard ICL and existing TV methods, we propose \textbf{DA4ICL}, an enhanced ICL framework combining a novel retrieval strategy and a refined TV anchoring mechanism. Our DA4ICL (see Fig.~\ref{fig1}) introduces two modules: (1) \textit{\textbf{Diverse and Contrastive Retrieval}} (DCR) to enrich demo sets along multiple complementary dimensions, and (2) \textit{\textbf{Projected Vector Anchoring}} (PVA) to anchor fine-grained, layer-wise TVs at demo-level $\to$ tokens. The DCR module mitigates the insufficient context width caused by conventional retrieval, while the PVA module realigns the injection granularity of TVs from test-centric to demo-centric anchoring, ensuring robust and context-deep adaptation for the AD detection task.

\subsection{Diverse and Contrastive Retrieval (DCR)}
\label{sec:dcr}
The DCR module aims to construct informative and contextually diverse demo sets in two sequential stages.

\paragraph{Stage 1: Main-Demo Set Construction (Width Enrichment).} 
For each test sample $d_{\mathrm{test}}$, we construct a \textit{main-demo} set by retrieving a pair of AD/HC demos under each of four complementary criteria: (i) semantic similarity, (ii) semantic dissimilarity, (iii) length similarity, and (iv) length dissimilarity. Let $\phi(d)$ represent the final-layer HS at the last $\to$ token of demo $d$, and $\ell(d)$ denote its sequence length. Formally, for each criterion $c\in\{\textit{sim}\phi,\textit{dis}\phi,\textit{sim}\ell,\textit{dis}\ell\}$, we select:
\begin{equation}
\small
\begin{aligned}
d^{c}_{+} &= \mathop{\arg\!\mathrm{ext}}_{(x,y)\in\mathcal{S},\,y=AD}\, f_c(d_{\mathrm{test}}, (x,y)), \\
d^{c}_{-} &= \mathop{\arg\!\mathrm{ext}}_{(x,y)\in\mathcal{S},\,y=HC}\, f_c(d_{\mathrm{test}}, (x,y)),
\end{aligned}
\end{equation}
where $\mathrm{ext}$ is $\max$ or $\min$ according to the criterion $c$, and $f_c$ denotes cosine similarity or length difference accordingly. The resulting \textit{main-demo} set $\mathcal{D}_{\mathrm{main}}$ comprises eight demos:
\begin{equation}
\small
    \mathcal{D}_{\mathrm{main}} = \{d^{c}_{+}, d^{c}_{-}\mid c\in\{\textit{sim}\phi,\textit{dis}\phi,\textit{sim}\ell,\textit{dis}\ell\}\}.
\end{equation}

\paragraph{Stage 2: Sub-Demo Set Construction (Depth Enrichment).}
To deepen and enrich the context around each \textit{main-demo} $d_i=(x_i,y_i)$ ($d_i\in \mathcal{D}_{\mathrm{main}}$), we further retrieve a set of 8 \textit{sub-demos} using the same four criteria but centered on $d_i$ itself. The resulting augmented sequence $\mathrm{Seq}_{i}$ for each main demo $d_i$ is:
\begin{equation}
\label{subdemo}
\small
\mathrm{Seq}_{i}^{sub} = \{x_{i,1}\!\to y_{i,1},\dots,x_{i,8}\!\to y_{i,8}, x_i\!\to\},
\end{equation}
where $\{(x_{i,j},y_{i,j})\}_{j=1}^{8}$ are the \textit{sub-demo} set (pre-computed) providing rich contrast and contextual detail. Crucially, each \textit{main-demo} serves as its own pseudo-sample for subsequent TV extraction, circumventing external pseudo-sample selection and enhancing representation consistency.

\subsection{Projected Vector Anchoring (PVA)}
\label{sec:pva}

Existing TV injection methods suffer from two key limitations: (1) \textit{they inject at only the test sample’s $\to$ token (test-centric), neglecting distributed demo cues}, and (2) \textit{they use addition or replacement, which distorts both the direction and magnitude of HSs, risking semantic misalignment}. To address these issues, we propose Projected Vector Anchoring (PVA), a demo-centric, layer-wise, and projection-based mechanism that ensures every demo anchor mostly contributes to the LLM’s reasoning at all depths.

\paragraph{Task Vector Extraction.}
In decoder-only LLMs, the HS at the last token and last layer overwhelmingly determines next-token prediction, due to the LLM’s architecture. Signals injected at earlier positions can easily be washed out during forward propagation. 

Therefore, to ensure that every demo anchor robustly contributes to the test prediction, we anchor TVs at every layer, making their influence persistent and cumulative throughout the residual stream. For each main demo $d_i$, we expand it with \textit{sub-demo} retrieval (see Eq.~\ref{subdemo}), and extract the HS at its $\to$ token $t_i$ of all layers:
\begin{equation}
\small
\mathbf{v}i^{(\ell)} = \mathbf{h}_{i,t_i}^{(\ell)},\quad \ell=1,\dots,L,
\end{equation}
where tokens in the $t_i-1$ and $t_i$-th position of $\mathrm{Seq}_{i}^{sub}$ are $\{x_i, \to\}$, where $\{x_i, \to\} \in \mathrm{Seq}_i^{sub}, x_i \in \mathcal{D}_{\mathrm{main}}$.

\paragraph{Projected and Layer-wise Anchoring.}
At inference, for each \textit{main-demo} $d_i$ and each layer $\ell$, we refine its HS by projecting the extracted TV onto the original representation, rather than naive addition or replacement, to ensure semantic alignment and robust adaptation.

First, we normalize the extracted TV to match the scale of the layer’s typical HSs:
\begin{equation}
\small
\bar v_i^{(\ell)} = \mu^{(\ell)} \cdot \frac{v_i^{(\ell)}}{|v_i^{(\ell)}|_2}
\end{equation}
where $\mu^{(\ell)}$ is the average $\ell_2$-norm of HSs at layer $\ell$.
This step ensures the injected signal is calibrated to the expected scale of LLM HSs.
Next, we compute the projection of the normalized TV onto the original HS direction:
\begin{equation}
\small
p_i^{(\ell)} = \frac{\langle \bar v_i^{(\ell)},h_{i,t_i}^{(\ell)}\rangle}{|h_{i,t_i}^{(\ell)}|_2^2+\epsilon} \cdot h_{i,t_i}^{(\ell)}
\end{equation}
This operation preserves the original semantic direction of the \textit{main-demo} anchor, modulating only its magnitude, and thereby avoids distorting semantic.
Finally, we update the HS with a layer-specific scaling factor:
\begin{equation}
h_{i,t_i}^{(\ell)\prime} = h_{i,t_i}^{(\ell)} + \gamma^{(\ell)} \cdot p_i^{(\ell)}.
\end{equation}
This flexible, projection-based anchoring ensures that demo-specific task information is injected in a manner that is both fine-grained and semantically consistent, leading to more stable and interpretable adaptation. More detailed mechanisms are listed in \textbf{Appendix: C}.

\begin{table*}[t]
\centering
\small
\setlength{\tabcolsep}{1mm}
\begin{tabular}{l c c | c c c c c c c c}
\toprule
\textbf{Dataset} & \textbf{Metric} (\%) & \(N\) 
  & ICL\(_{\mathrm{Van}}\) & ICL\(_{\mathrm{Sem}}\) & ICL\(_{\mathrm{Ens}}\) 
  & TV\(_{\mathrm{Van}}^{\mathrm{Add}}\) & TV\(_{\mathrm{Sem}}^{\mathrm{Add}}\) 
  & TV\(_{\mathrm{Van}}^{\mathrm{Rep}}\) & TV\(_{\mathrm{Sem}}^{\mathrm{Rep}}\) 
  & Ours \\
\midrule
\multirow{10}{*}{\textbf{Test}}  
 & \multirow{5}{*}{F1} 
   & 0 & 51.50$\pm{7.34}$ & -           & 54.71$\pm{7.20}$ & -           & -           & -           & -           & - \\
 &                      & 1 & 69.33$\pm{4.22}$ & 68.93$\pm{3.88}$ & 64.43$\pm{4.62}$ & 51.23$\pm{6.62}$ & 52.28$\pm{8.33}$ & 67.11$\pm{3.12}$ & 65.00$\pm{4.94}$ & - \\
 &                      & 2 & 68.29$\pm{3.71}$ & 68.70$\pm{5.14}$ & 71.42$\pm{3.71}$ & 54.60$\pm{5.77}$ & 56.02$\pm{8.84}$ & 68.40$\pm{4.27}$ & 64.46$\pm{2.86}$ & - \\
 &                      & 3 & 70.66$\pm{1.83}$ & 73.71$\pm{4.73}$ & 70.20$\pm{5.24}$ & 51.04$\pm{6.17}$ & 51.44$\pm{7.95}$ & 66.94$\pm{4.72}$ & 69.68$\pm{5.40}$ & - \\
 &                      & 4 & 72.13$\pm{4.15}$ & 74.20$\pm{3.69}$ & \underline{75.32$\pm{4.08}$} & 54.47$\pm{5.21}$ & 56.31$\pm{7.99}$ & 66.97$\pm{3.96}$ & 64.15$\pm{4.65}$ & \textbf{86.11$^{\dag}\pm{1.92}$} \\
\cmidrule(r){2-11}
 & \multirow{5}{*}{Acc.} 
   & 0 & 57.71$\pm{6.46}$ & -           & 59.58$\pm{5.76}$ & -           & -           & -           & -           & - \\
 &                      & 1 & 67.50$\pm{5.12}$ & 66.46$\pm{4.00}$ & 62.50$\pm{5.19}$ & 54.58$\pm{5.88}$ & 57.08$\pm{5.12}$ & 62.92$\pm{3.46}$ & 61.46$\pm{5.76}$ & - \\
 &                      & 2 & 66.67$\pm{3.36}$ & 66.88$\pm{5.39}$ & 70.21$\pm{3.73}$ & 58.13$\pm{5.22}$ & 61.04$\pm{6.72}$ & 65.21$\pm{4.66}$ & 59.17$\pm{2.83}$ & - \\
 &                      & 3 & 68.33$\pm{2.43}$ & 71.25$\pm{5.65}$ & 68.96$\pm{5.31}$ & 55.62$\pm{6.79}$ & 57.08$\pm{4.86}$ & 62.08$\pm{4.82}$ & 65.42$\pm{5.45}$ & - \\
 &                      & 4 & 70.00$\pm{4.08}$ & 71.04$\pm{4.32}$ & \underline{71.46$\pm{5.44}$} & 56.88$\pm{4.85}$ & 59.79$\pm{6.85}$ & 62.92$\pm{5.34}$ & 59.38$\pm{4.95}$ & \textbf{85.83$^{\dag}\pm{1.91}$} \\
\midrule
\multirow{10}{*}{\textbf{Lu}}  
 & \multirow{5}{*}{F1} 
   & 0 & 63.41$\pm{7.24}$ & -           & 63.59$\pm{5.59}$ & -           & -           & -           & -           & - \\
 &                      & 1 & 76.45$\pm{3.03}$ & 77.89$\pm{2.69}$ & 77.71$\pm{3.99}$ & 61.10$\pm{7.41}$ & 64.79$\pm{6.62}$ & 75.31$\pm{4.16}$ & 74.10$\pm{3.48}$ & - \\
 &                      & 2 & 79.31$\pm{2.86}$ & 78.01$\pm{2.65}$ & 78.91$\pm{3.12}$ & 62.12$\pm{6.34}$ & 64.54$\pm{4.53}$ & 76.06$\pm{1.71}$ & 74.51$\pm{3.19}$ & - \\
 &                      & 3 & 79.83$\pm{4.10}$ & 78.81$\pm{2.54}$ & 80.44$\pm{3.24}$ & 58.42$\pm{6.72}$ & 61.55$\pm{5.31}$ & 76.75$\pm{3.43}$ & 73.77$\pm{3.54}$ & - \\
 &                      & 4 & 79.53$\pm{3.58}$ & \underline{82.02$\pm{2.11}$} & 79.88$\pm{3.28}$ & 62.06$\pm{7.81}$ & 61.91$\pm{5.65}$ & 74.46$\pm{4.28}$ & 76.54$\pm{3.46}$ & \textbf{86.12$^{\dag}\pm{1.73}$} \\
\cmidrule(r){2-11}
 & \multirow{5}{*}{Acc.} 
   & 0 & 55.71$\pm{8.05}$ & -           & 56.19$\pm{5.65}$ & -           & -           & -           & -           & - \\
 &                      & 1 & 66.90$\pm{4.05}$ & 69.52$\pm{3.33}$ & 68.57$\pm{5.19}$ & 54.76$\pm{6.82}$ & 59.29$\pm{6.43}$ & 65.00$\pm{5.43}$ & 63.57$\pm{3.99}$ & - \\
 &                      & 2 & 70.95$\pm{4.10}$ & 69.05$\pm{3.53}$ & 70.00$\pm{4.29}$ & 55.48$\pm{6.03}$ & 57.38$\pm{4.32}$ & 66.43$\pm{2.49}$ & 63.57$\pm{4.27}$ & - \\
 &                      & 3 & 71.19$\pm{6.25}$ & 68.81$\pm{3.44}$ & 72.14$\pm{4.27}$ & 53.10$\pm{5.84}$ & 54.52$\pm{4.70}$ & 67.14$\pm{4.86}$ & 62.86$\pm{4.54}$ & - \\
 &                      & 4 & 70.95$\pm{5.08}$ & \underline{73.33$\pm{3.50}$} & 70.95$\pm{4.97}$ & 55.48$\pm{8.11}$ & 55.71$\pm{5.65}$ & 63.57$\pm{5.33}$ & 66.43$\pm{5.05}$ & \textbf{80.95$^{\dag}\pm{2.51}$} \\
\midrule
\multirow{10}{*}{\textbf{Pitt}}  
 & \multirow{5}{*}{F1} 
   & 0 & 55.23$\pm{2.00}$ & -           & 56.30$\pm{2.67}$ & -           & -           & -           & -           & - \\
 &                      & 1 & 67.18$\pm{1.11}$ & 67.59$\pm{1.21}$ & 67.41$\pm{1.89}$ & 55.33$\pm{2.24}$ & 54.71$\pm{1.72}$ & 66.92$\pm{1.21}$ & 67.00$\pm{0.72}$ & - \\
 &                      & 2 & 67.74$\pm{1.35}$ & 68.18$\pm{1.83}$ & 69.11$\pm{1.40}$ & 53.86$\pm{2.35}$ & 54.31$\pm{2.38}$ & 66.31$\pm{1.15}$ & 67.63$\pm{1.63}$ & - \\
 &                      & 3 & 68.81$\pm{1.40}$ & 71.70$\pm{0.83}$ & 69.44$\pm{0.93}$ & 55.10$\pm{2.33}$ & 54.22$\pm{2.65}$ & 67.10$\pm{1.44}$ & 66.99$\pm{1.23}$ & - \\
 &                      & 4 & 70.24$\pm{1.12}$ & \underline{73.79$\pm{1.21}$} & 70.66$\pm{1.07}$ & 53.45$\pm{2.44}$ & 52.60$\pm{1.80}$ & 67.45$\pm{1.02}$ & 67.64$\pm{1.39}$ & \textbf{80.23$^{\dag}\pm{0.42}$} \\
\cmidrule(r){2-11}
 & \multirow{5}{*}{Acc.} 
   & 0 & 55.72$\pm{1.96}$ & -           & 56.92$\pm{2.21}$ & -           & -           & -           & -           & - \\
 &                      & 1 & 63.33$\pm{1.12}$ & 64.34$\pm{1.21}$ & 63.42$\pm{1.83}$ & 55.06$\pm{1.62}$ & 54.39$\pm{1.62}$ & 59.85$\pm{1.43}$ & 59.64$\pm{0.91}$ & - \\
 &                      & 2 & 64.63$\pm{1.15}$ & 65.39$\pm{1.79}$ & 66.01$\pm{1.59}$ & 53.92$\pm{2.11}$ & 53.83$\pm{2.44}$ & 59.03$\pm{1.27}$ & 60.55$\pm{1.82}$ & - \\
 &                      & 3 & 65.43$\pm{1.55}$ & 68.12$\pm{0.93}$ & 66.25$\pm{0.95}$ & 54.35$\pm{1.37}$ & 53.77$\pm{2.31}$ & 59.67$\pm{1.71}$ & 59.82$\pm{1.55}$ & - \\
 &                      & 4 & 66.48$\pm{1.20}$ & \underline{69.71$\pm{1.13}$} & 67.41$\pm{1.10}$ & 53.57$\pm{2.09}$ & 53.01$\pm{1.71}$ & 60.27$\pm{1.06}$ & 60.44$\pm{1.38}$ & \textbf{79.42$^{\dag}\pm{0.39}$} \\
\bottomrule
\end{tabular}
\caption{Mean $\pm$ std of F1-score and Accuracy over 10 runs for ICL (Van, Sem, Ens), TV (Van\textsuperscript{Add}, Sem\textsuperscript{Add}, Van\textsuperscript{Rep}, Sem\textsuperscript{Rep}), and DA4ICL (Ours) on three AD detection datasets (Test, Lu, Pitt) with varying demo counts (N). Bold entries denote the best result, and underlined entries denote the second-best. Significance is shown with $\dag$ for DA4ICL compared to second-best baselines. Statistical significance was measured with a paired t-test ($p < 0.005$) and a Wilcoxon signed-rank test ($p < 0.01$).}
\label{tab:main_results_transposed}
\end{table*}

\section{Experiments}

\subsection{Experimental Setup}

Our experiments are structured to progressively analyze how DA4ICL overcomes the key challenges in AD detection, which are, limited task perception, insufficient demo diversity, and mismatched injection granularity.

We begin by benchmarking DA4ICL against ICL and TV baselines under varied retrieval and inference settings, revealing the performance saturation of ICL and the ineffectiveness of test-centric TV injection. We then isolate the role of demo diversity by applying our retrieval strategy (DCR) to both DA4ICL and existing ICL/TV pipelines, confirming its universal benefit for ICL but limited utility for conventional TV methods. Finally, we investigate why TV fails to leverage DCR, demonstrating that our projection-based, multi-layer anchoring (PVA) is essential to effectively inject fine-grained task signals. \textit{\textbf{Together, these experiments explain both the failure modes of previous methods and the mechanisms behind DA4ICL's improvements.}}

\paragraph{Datasets.}  
We perform experiments on three widely-adopted AD corpora: the \textit{ADReSS Challenge dataset}~\cite{luz2020alzheimer}, the \textit{Lu corpus}~\cite{lu}, and the \textit{Pitt corpus}~\cite{pitt}. The \textit{ADReSS Train} split contains 54 AD and 54 HC (\textit{Test} split: 24 vs. 24), providing a balanced in-distribution benchmark. In contrast, the \textit{Lu corpus} comprises 15 AD vs. 27 HC, and the \textit{Pitt corpus} comprises 243 AD vs. 306 HC, introducing pronounced class imbalance. For our  experiments, every demo is drawn exclusively from ADReSS Train split, with the rest corpora strictly held out for evaluation. All transcripts follow the standard CHAT protocol~\cite{macwhinney2000childes}, ensuring consistent preprocessing and enabling direct cross-corpus comparison. More details can be found in \textbf{Appendix: A}.

\paragraph{Baselines.}
We compare our method comprehensively against two main families of strong LLM-based baselines: \textbf{ICL} methods and \textbf{TV} methods. Within the ICL family, we evaluate: (1) \textit{Vanilla ICL}, which randomly selects demonstrations from the support set without considering relevance, (2) \textit{Semantic ICL}~\cite{demo_sem_1}, which retrieves demos semantically closest to the test sample, measured via cosine similarity in latent space, and (3) \textit{Ensemble ICL}~\cite{ensemble_1}, which aggregates predictions from multiple independently sampled or retrieved demonstration sets through majority voting, aiming to enhance prediction robustness. Collectively, these ICL variants examine both the construction of demonstration contexts and the inference mechanisms within conventional ICL paradigms. For the TV family, we evaluate methods that directly manipulate HSs to inject latent task representations. Specifically, we test two core injection approaches: (4) \textit{Replace TV}~\cite{tv_first}, which replaces the HS at final $\to$ position directly with a TV derived from demos, and (5) \textit{Add TV}~\cite{tv_icv}, which injects the TV into the original HS via addition. Both methods consider two variants in demos retrieval for extraction: random retrieval and semantic similarity-based retrieval. Thus, these TV baselines thoroughly explore the mechanisms (replacement vs. addition) and the content (random vs. semantic) of latent task signal injection, allowing precise comparisons to our proposed demo-centric anchoring strategy.

\paragraph{Implementation Details.}  
We implement all methods using the \texttt{Llama3.1-8B-instruct} model as the backbone to ensure consistent and fair comparisons, and conduct experiments on a single NVIDIA Quadro RTX 8000 (48GB GPU). To stabilize outputs, we set temperature of $0.1$ and top-$k$ of $50$ across all experiments. For demo retrieval, we uniformly sample demos in pairs (AD/HC per pair) and evaluate performance under varying demo pair counts. For TV baselines, TVs are consistently extracted and injected at the final HS layer at the last token ($\to$) of the input sequence. The injection strength parameter is set as $0.5$ for Add TV. For DA4ICL, the $\gamma(\ell)$ is set as $1$ for $\ell\in[0,7]\cup[24,31]$, and $0.2$ for $\ell\in[8,23]$. More details are listed in \textbf{Appendix: D}.

\paragraph{Evaluation Metrics.}  
We evaluate performance comprehensively using accuracy and F1-score. All experimental results are averaged over 10 independent runs under identical conditions to ensure statistical robustness and reliability.

\subsection{Main Results}

Tab.~\ref{tab:main_results_transposed} presents the comparative results of DA4ICL and a comprehensive set of baselines, including vanilla, semantic, ensemble ICL, and two representative TV injection methods on three AD detection datasets.

\paragraph{LLM's task cognition is insufficient for AD detection.} The zero-shot results (ICL$_{Van}$, $N$=0) are consistently poor across all datasets (e.g., Acc 57.71\% on Test/Lu and 55.72\% on Pitt), directly confirming that LLMs lack the necessary task cognition for AD detection in the absence of demos.

\paragraph{Limited gains from current ICL strategies.} Increasing the $N$ or switching from random to semantic retrieval yields limited and fluctuated improvements. For instance, on Test, accuracy rises only from 67.50\% ($N=1$) to 70.00\% ($N=4$), from 70.00\% (vanilla, $N=4$) to 71.04\% (semantic, $N=4$), and F1 scores follow a similar trend. Notably, on the Lu dataset, both F1 scores and accuracy decrease when moving from ICL$_{Van}$ to ICL$_{Sem}$ ($N=2,3$), illustrating instability and lack of robustness. ICL$_{Ens}$ offers minor gains (Test F1/Acc: 75.32/71.46\%), further confirming the limited effect.

\paragraph{Conventional TV injection fails for AD adaptation.} Both addition and replacement TV methods consistently underperform ICL baselines, regardless of retrieval or injection strategy across dataset. For example, TV$_{Van}^{Add}$ and TV$_{Sem}^{Add}$ achieve only 56.31/54.47\% F1 and TV$_{Van}^{Rep}$ and TV$_{Sem}^{Rep}$ reaches 66.97/64.15\% F1 on Test, well below ICL, and with accuracy following the same trend. Importantly, semantic retrieval provides no consistent benefit. These findings confirm that single-layer, last-token TV injection fundamentally fails to support robust task adaptation in AD detection.

\paragraph{DA4ICL achieves substantial and stable improvements.} In contrast, DA4ICL delivers the highest F1 and accuracy across all datasets (Test: 86.11/85.83\%, Lu: 86.12/80.95\%, Pitt: 80.23/79.42\%), with the lowest variance among all methods. These results confirm that enriching demos through diverse, contrastive retrieval and demo-wise vector anchoring decisively overcomes the adaptation bottlenecks and granularity mismatches of prior ICL and TV paradigms, enabling reliable and robust AD detection.

\begin{figure*}[t]
\centering
\includegraphics[width=1\textwidth]{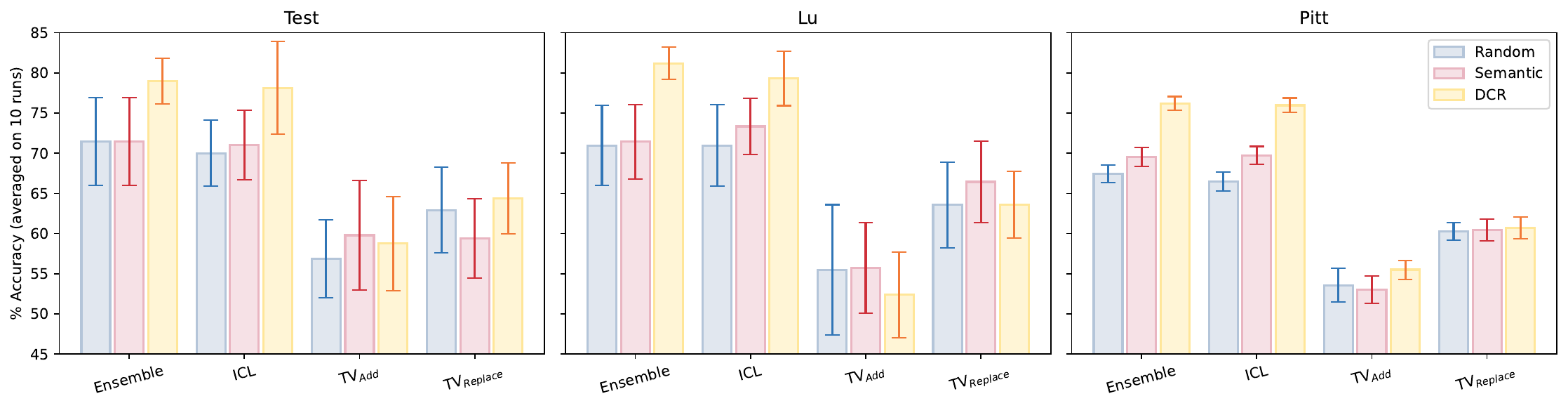}
\caption{Accuracy comparison after applying DCR to ICL and TV methods (mean $\pm$ std over 10 runs). DCR consistently improves ICL performance across all datasets, demonstrating the value of contrastive demo construction. However, DCR fails to enhance conventional TV methods, with performance remaining stable or degrading, highlighting the limitations of TV’s injection granularity and alignment. Specific results can be found in Appendix: B.} 
\label{fig5}                        
\end{figure*}

\begin{figure}[t]
\centering
 \includegraphics[height=7cm]{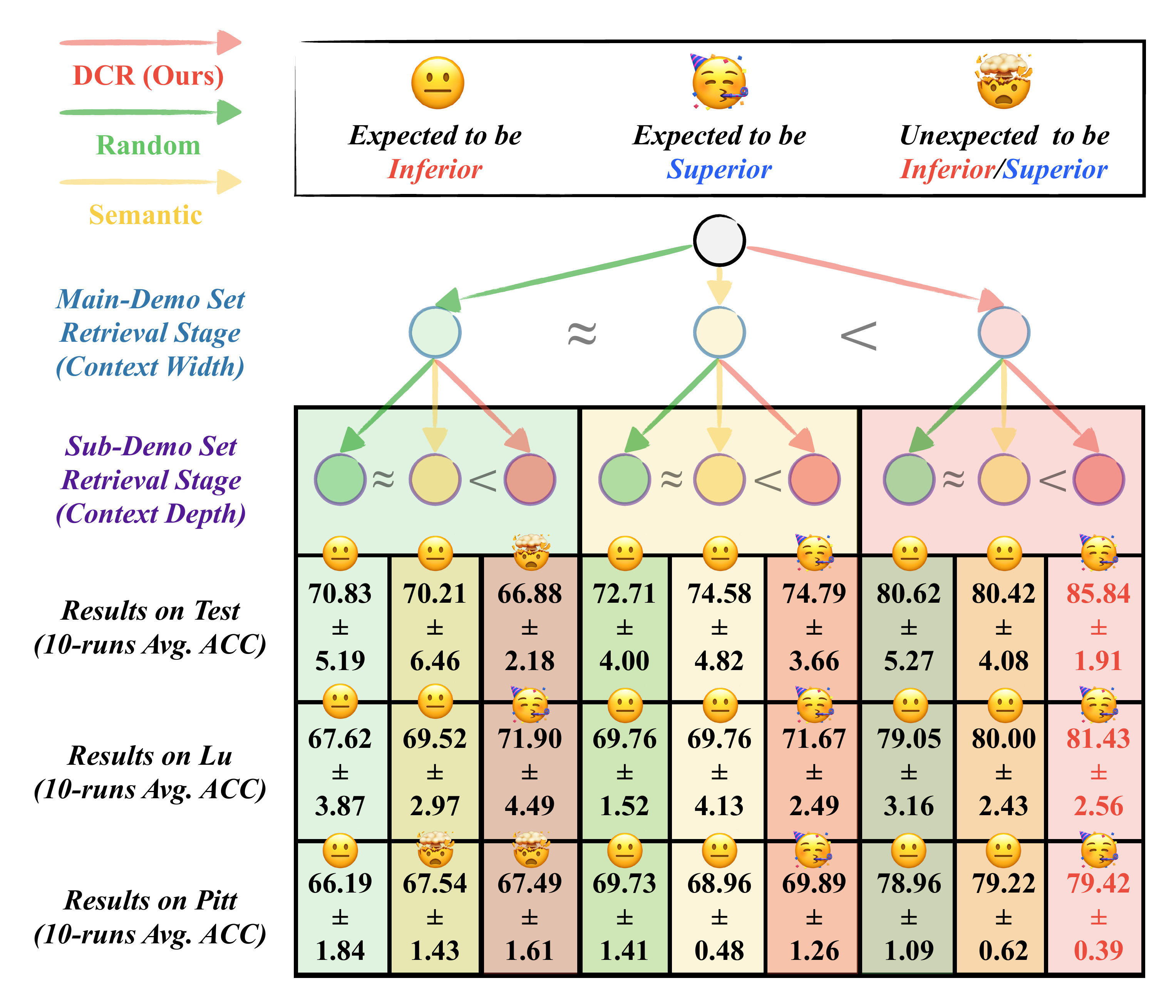}
\caption{Ablation over two-stage retrieval strategies. Diverse main-demo selection (context width) provides the largest gains, while sub-demo enrichment (context depth) offers further complementary improvements.} 
\label{fig6}
\end{figure}

\subsection{Dissecting Two-Stage Retrieval and the Generality of DCR}

We first examine the effectiveness of DCR from two perspectives: within our full DA4ICL framework and when integrated into existing ICL and TV baselines.

\paragraph{Context width is essential and depth provides complementary gains.}
Ablation of retrieval strategies (Fig.~\ref{fig6}) reveals that applying DCR to the \textit{main-demo} stage consistently yields the highest accuracy and lowest variance, regardless of the \textit{sub-demo} strategy. Removing DCR from the \textit{main-demo} stage, replacing it with random or semantic retrieval, leads to notable drops in performance, highlighting that context width (diverse and contrastive \textit{main-demos}) is a prerequisite for effective adaptation, while context depth (\textit{sub-demos}) provides complementary gains.

\paragraph{DCR generalizes to ICL, but fails to activate standard TV methods.}
We further apply DCR to ICL$_{Van}$, ICL$_{Ens}$, TV$_{Add}$, and TV$_{Rep}$ (Fig.~\ref{fig5}). Results show that DCR significantly boosts both standard and ensemble ICL across all datasets, confirming that contrastive demo construction improves robustness and generalization. However, such improvement fails to transfer to conventional TV methods, performance remains unchanged or even degraded. This suggests that existing TV methods cannot effectively absorb the diverse, fine-grained signals DCR provides, due to their injection granularity and alignment limitations.

\begin{table}[ht]
\centering
\small
\setlength{\tabcolsep}{1mm}
\begin{tabular}{l|cccc}
\toprule
\multirow{2}{*}{\textbf{Dataset}} &  \multicolumn{4}{c}{\textbf{Acc} (\%)} \\
    & Addition & Replacement & w/o injection & PVA\\
\midrule
Test  & 77.50$\pm{4.04}$ & 69.38$\pm{2.95}$& 77.29$\pm{4.41}$ & \textbf{85.83$\pm{1.91}$} \\
Lu    & 77.62$\pm{3.23}$ & 75.71$\pm{1.78}$ & 77.86$\pm{2.83}$ & \textbf{81.43$\pm{2.56}$} \\
Pitt  & 76.41$\pm{0.91}$ & 72.82$\pm{1.00}$ & 75.94$\pm{1.21}$ & \textbf{79.42$\pm{0.39}$} \\
\bottomrule
\end{tabular}
\caption{Ablation on anchoring (demo-centric) methods. Projection-based PVA achieves the best performance, outperforming addition, replacement, and removal variants across all datasets.}
\label{tab:ablation_on_PVA}
\end{table}

\subsection{Why Direct TV Injection Fails and PVA Matters}

To understand why standard TV methods struggle, we further dissect their injection strategy and contrast them with our proposed PVA module.

\paragraph{Test-centric injection with coarse granularity undermines adaptation.}
Standard TV methods inject TVs only at test sample’s $\to$ token and single layer. Such test-centric, single-token and single-layer injection discards distributed demo cues and fails to propagate fine-grained distinctions. As shown in Fig.~\ref{fig5}, even when empowered with DCR, these methods do not improve and often performing worse than simpler ICL strategies. This validates the need for demo-centric, multi-layer anchoring.

\paragraph{PVA preserves semantic alignment and achieves fine-grained control.}
Ablation results in Tab.~\ref{tab:ablation_on_PVA} confirm that PVA’s projection-based, layer-wise anchoring substantially outperforms naive alternatives. These results show that direct addition or replacement disrupts the original semantic direction of the HSs, while PVA preserves alignment and allows stable, context-sensitive adaptation. Notably, PVA can amplify both helpful and harmful cues of the  set, emphasizing the importance of robust retrieval strategy like DCR.

\section{Conclusion}
We revisit the core limitations of ICL and TV methods for AD detection, showing that both demo context narrowness and misaligned vector injection hinder task perception. DA4ICL addresses these challenges by integrating diverse and contrastive retrieval strategy with demo-centric, projection-based TV anchoring, jointly expanding context width and depth while preserving semantic alignment. Experiments on three AD datasets demonstrate substantial and stable improvements over previous methods. We believe this demo-centric anchoring paradigm offers a promising foundation for fine-grained, low-resource  and OOD task adaptation with LLMs.

\section{Acknowledgments}
This work was supported by the Key Research and Development Project of Hunan Province (No. 2025JK2119), Foundation of NUDT (HQKYZH2025KD004), the Leading Science and Technology Innovation Talents Program of Furong Project (2025RC1048).

\bibliography{aaai2026}
\appendix
\section{Appendix}
\begin{figure}[!ht]
    \centering
    \includegraphics[width=\linewidth]{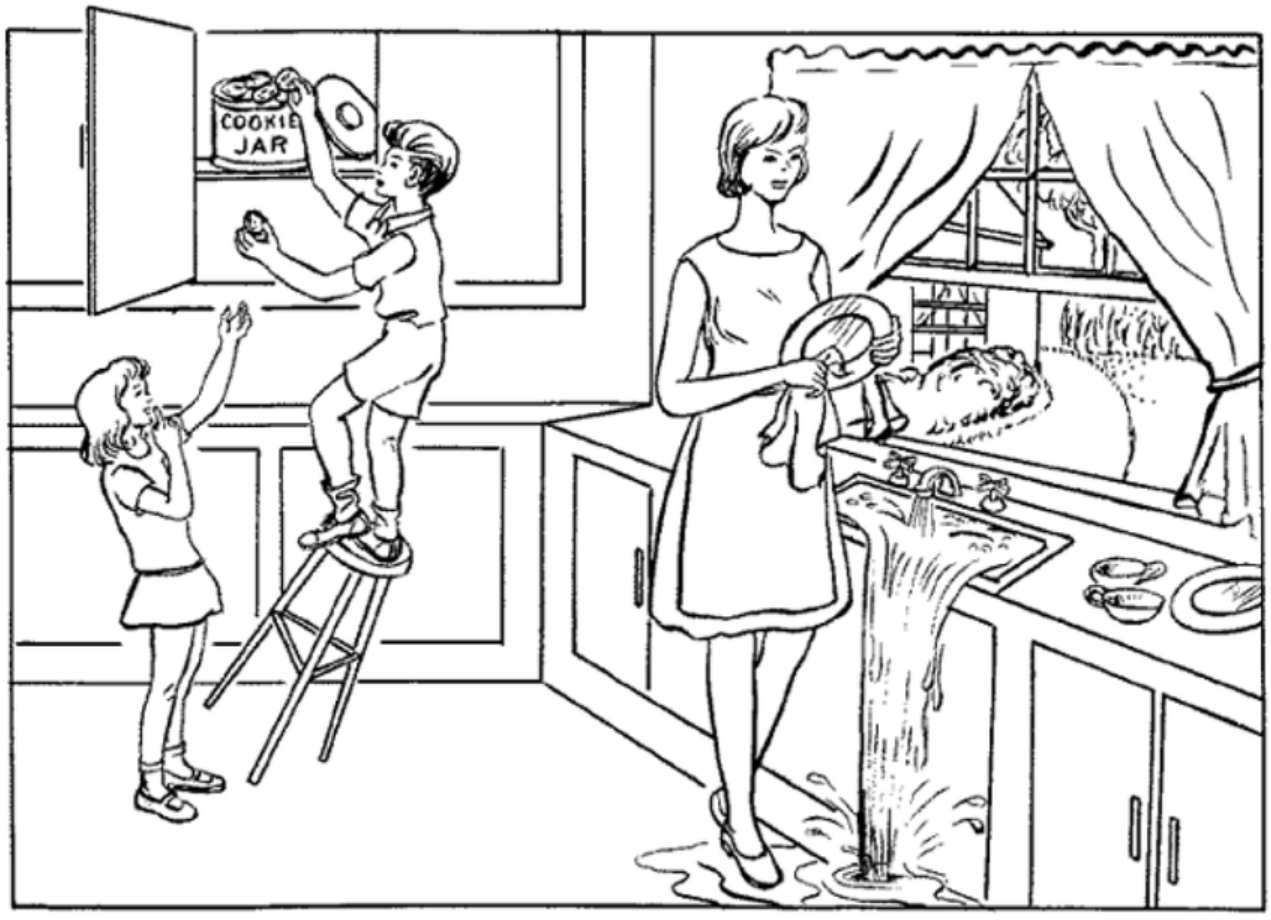}
    \caption{Image for the Cookie Theft picture description task. All participants are asked to describe the details and actions present in this scene.}
    \label{fig:cookie_theft}
\end{figure}

\noindent\textbf{Appendix Overview.}
This appendix complements the technical content of our paper with task specifications, implementation details, mechanistic analyses, and additional experiments, following the structure below:
\paragraph{Appendix A. \textit{Task and Dataset Details.}} defines the task (\textit{A.1 Task Definition: Alzheimer's Disease Detection from Narrative Speech}), summarizes datasets, demographics and licenses (\textit{A.2 Datasets, Demographics and Licenses}), details preprocessing (\textit{A.3 Preprocessing}), and highlights key challenges (\textit{A.4 Task Challenges}). Correlated figures and tables are  Fig.~\ref{fig:cookie_theft}, Tab.~\ref{table:dataset}.

\paragraph{Appendix B. \textit{Supplementary Implementation Details.}} documents the backbone and hardware (\textit{B.1 Details of Backbone Model}: \textit{Model and Hardware}, \textit{Model Scale Considerations}); demo construction (\textit{B.2 Details of Demo Construction}: \textit{Prompt Construction}, \textit{Demo Retrieval and Construction}, \textit{Hyperparameters and Inference}); and experimental settings (\textit{B.3 Details of Experimental Settings}: \textit{Statistical Analysis}, \textit{Evaluation Metrics}). Correlated figures and tables are Fig.~\ref{fig:layerwise_pca}, Tab.~\ref{tab:prompt_template}.

\paragraph{Appendix C. \textit{Supplementary Mechanistic Analysis.}} formalizes ICL in the residual-stream view (\textit{C.1 In-Context Learning (ICL): Residual Stream Formulation}: \textit{Residual Stream in Transformers}, \textit{Generation Flow in ICL}, \textit{Final Token as Contextual Anchor}, \textit{Layer-wise Information Refinement}); specifies TV extraction/injection (\textit{C.2 Task Vector (TV) Method: Extraction and Injection Mechanism}: \textit{Task Vector Extraction}, \textit{Single-point Task Vector Injection}, \textit{Vertical Propagation and Horizontal Limitation}); visualizes design limits (\textit{C.3 Visualizing Limitations of TV Injection}); and details our diverse/contrastive retrieval (\textit{C.4 Demo Construction: Diverse and Contrastive Retrieval}: \textit{Stage~1: Main-Demo Set Construction}, \textit{Remarks}). Correlated figures and tables are Fig.~\ref{fig:tv_injection_effects}, Fig.~\ref{fig:concept_format}, Tab.~\ref{tab:hyperparam}.

\paragraph{Appendix D. \textit{Supplementary Experiments.}} reports ablations for PVA (\textit{D.1 Ablation Results for PVA Module}), full metrics across datasets and shots (\textit{D.2 Full Evaluation Metrics}), supervised baselines and LoRA configurations (\textit{D.3 Comparison of Supervised Baselines}: \textit{Fine-tuning Methods}, \textit{Detailed Information of LoRA}, \textit{Hyperparameters}, \textit{Classification Task Setup}, \textit{Generation Task Setup}), efficiency/resource measurements on 48\,GB/80\,GB GPUs (\textit{D.4 Efficiency and Resource Analysis}), and a quantitative–qualitative error analysis of Top-2 runs (\textit{D.5 Error Analysis}: \textit{Overall Trends}, \textit{Error Direction Asymmetry}, \textit{Qualitative findings}). Correlated figures and tables are Fig.~\ref{fig:error_analysis}, Tab.~\ref{tab:pva_ablation_full}, Tab.~\ref{tab:lora_hyperparameters}, Tab.~\ref{tab:supervised_bibm_combined}, Tab.~\ref{tab:efficiency_48g}, Tab.~\ref{tab:efficiency_80g}, Tab.~\ref{tab:top2_error_details}, Tab.~\ref{tab:lora_classification}, Tab.~\ref{tab:lora_generation}, Tab.~\ref{tab:main_results_full}.


\begin{table*}[!ht]
  \centering
  \begin{tabular}{l|cc|cc}
    \toprule
    \multirow{2}{*}{Dataset} &
      \multicolumn{2}{c|}{Alzheimer's Disease} &
      \multicolumn{2}{c}{Healthy Control} \\
     & Age (mean$\pm$std) & Gender (M/F) & Age (mean$\pm$std) & Gender (M/F) \\
    \midrule
    Train & 66.4 $\pm$ 6.4 & 24 / 30 & 66.8 $\pm$ 6.6 & 24 / 30 \\
    Test  & 66.1 $\pm$ 6.9 & 11 / 13 & 66.1 $\pm$ 7.3 & 11 / 13 \\
    Lu    & 73.9 $\pm$ 8.6 &  7 /  8 & 79.2 $\pm$ 8.9 & 12 / 15 \\
    Pitt  & 64.9 $\pm$ 7.7 & 89 / 154 & 71.4 $\pm$ 8.5 & 117 / 189 \\
    \bottomrule
  \end{tabular}
  \caption{Demographic statistics of AD and HC groups in the ADReSS (Train/Test), Lu, and Pitt corpora.}
  \label{table:dataset}
\end{table*}

\begin{table*}[!ht]
\centering
\begin{tabular}{lp{13cm}}
\toprule
\textbf{Prompt Component} & \textbf{Content Example} \\
\midrule
System Prompt&
\texttt{You are an experienced clinician specializing in dementia care. You will analyze descriptions provided by subjects who are describing a scene from the "Cookie Theft" picture, used in the Boston Diagnostic Aphasia Examination. In this task, subjects are asked to describe everything they see in the picture, which depicts a chaotic kitchen scene ...   \linebreak Your task is to analyze the subject's description to determine whether it indicates [HEALTHY CONTROL] mental condition or [Alzheimer's Disease] mental condition.\linebreak Give your answer (Start with [HEALTHY CONTROL] or Start with [Alzheimer's Disease]) first.} \\
\midrule
Demo Example 1 & 
\texttt{Here is the example: \linebreak \#\#Text:<The mother is washing the dishes ...> \linebreak\#\#Answer: The condition of this description is -> HEALTHY CONTROL} \\
\midrule
Demo Example 2 & 
\texttt{Here is the example: \linebreak \#\#Text:<The boy is standing on a stool to reach ...> \linebreak \#\#Answer: The condition of this description is -> HEALTHY CONTROL} \\
\midrule
Test Query & 
\texttt{Now you should analyze the testing subject's description: \linebreak \#\#Text:<Subject's description here> \linebreak \#\#Answer: I think The condition of this description will be ->} \\
\bottomrule
\end{tabular}
\caption{Prompt construction template for ICL and TV methods. The system prompt provides clinical context. Each demo example presents a subject's Cookie Theft description and its label. The test query asks the model to analyze a new description and output the predicted condition. All prompts are in English.}
\label{tab:prompt_template}
\end{table*}


\subsection{A. Task and Dataset Details}

\subsubsection{A.1 Task Definition: Alzheimer’s Disease Detection from Narrative Speech}

The primary task evaluated in this study is Alzheimer’s disease detection from narrative speech transcripts, specifically through the Cookie Theft picture description protocol. In this task, participants are shown a standard image (see Fig.~\ref{fig:cookie_theft}) depicting a kitchen scene with several salient elements. Each participant is instructed to describe, in as much detail as possible, everything occurring in the image. Their spoken descriptions are transcribed and subsequently used as input for automatic AD detection.

The task is to classify each transcript as originating from an individual with Alzheimer’s disease (AD) or from a healthy control (HC). This presents a challenging scenario due to the high semantic similarity of transcripts (as all subjects describe the same scene) and the subtle linguistic differences associated with early-stage AD.

\subsubsection{A.2 Datasets, Demographics and Licenses}

Experiments are conducted on three benchmark corpora derived from DementiaBank: ADReSS (Train/Test), Lu, and Pitt. Tab.~\ref{table:dataset} summarizes the demographic information for each split.

\textbf{ADReSS (Train/Test):} Age- and gender-balanced splits, ensuring comparability between AD and HC groups.  
\textbf{Lu:} Contains older participants with a modest gender imbalance, serving as an out-of-distribution evaluation scenario.  
\textbf{Pitt:} The largest corpus, with greater demographic skew and the most challenging OOD setting.

All datasets are sourced from DementiaBank (Train, Test, Lu, Pitt). The DementiaBank is a shared database of multimedia interactions for the study of communication in dementia, a sub database of TalkBank. Train/test splits are strictly separated, and no overlap exists. The use of TalkBank data is governed by the Creative Commons CC BY-NC-SA 3.0 copyright license. and subject to additional requirements:
\begin{itemize}
    \item Researchers must abide by the TalkBank Code of Ethics and maintain confidentiality NIH Certificate of Confidentiality.
    \item Access to password-protected corpora is limited to authorized academic members, and students must be supervised by faculty.
    \item Raw data cannot be redistributed or uploaded to external platforms.
\end{itemize}
\textbf{Note:} All processed results in this paper are based on data access compliant with DementiaBank policies. Data cannot be made public due to license restrictions but can be obtained by qualified researchers through DementiaBank as above.

\subsubsection{A.3 Preprocessing}

All audio recordings are transcribed to text. Preprocessing includes removal of non-speech markers, normalization of punctuation, and lowercasing. Only transcripts corresponding to the Cookie Theft description are retained. Speaker and interviewer markers are excluded.

\subsubsection{A.4 Task Challenges}

Challenges include limited data size and pre-training exposure (due to privacy and recruitment constraints) and semantic homogeneity (since all participants describe the same image). These characteristics make the AD detection an out-of-distribution (OOD) task, and hinder its adaptation and generalization ability.

\begin{figure}[!ht]
\centering
\includegraphics[width=\linewidth]{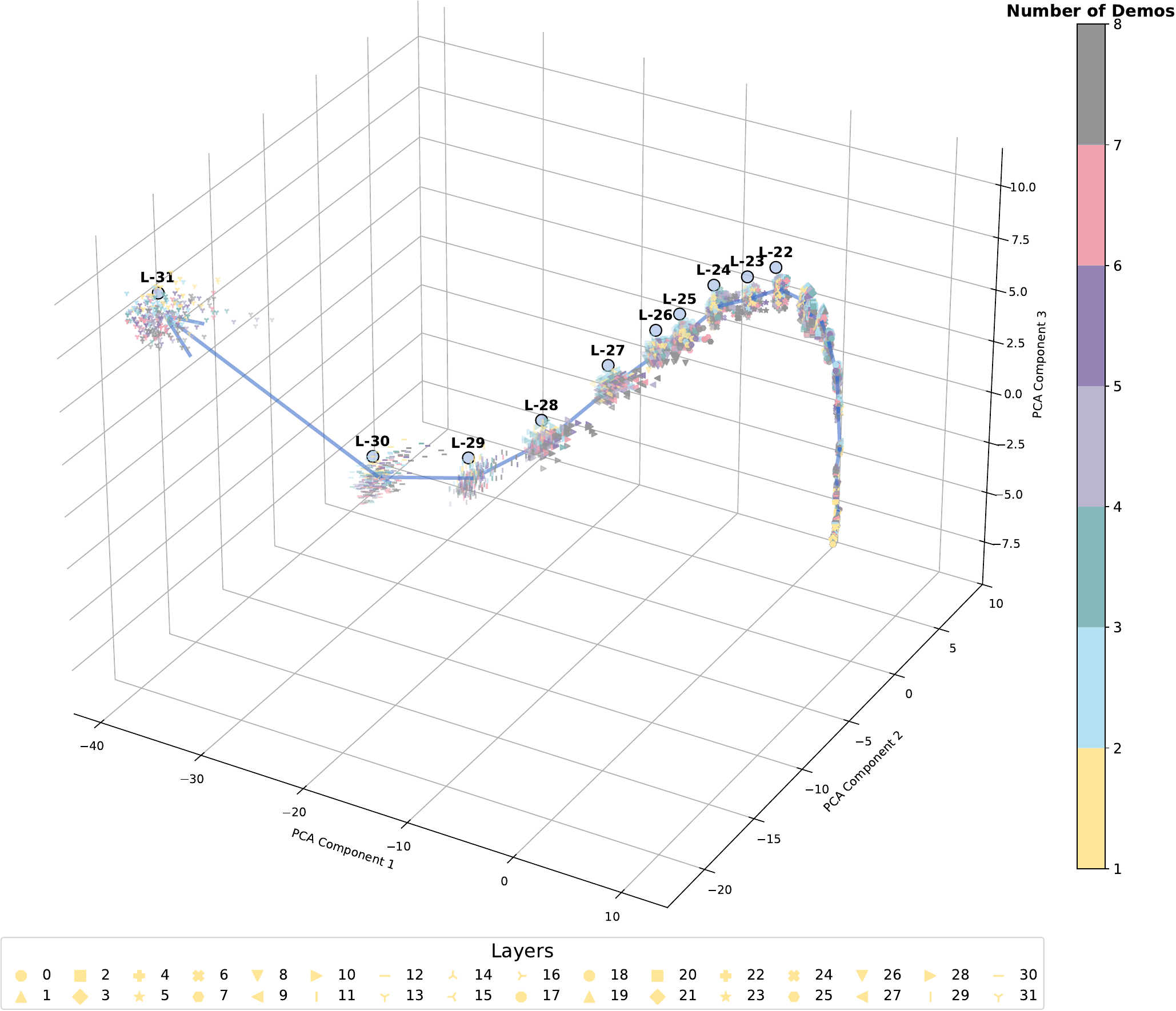}
\caption{Layer-wise PCA visualization of demo separator token representations from shallow (right) to deep (left) layers. Clustering by class emerges in higher layers, illustrating progressive semantic refinement.}
\label{fig:layerwise_pca}
\end{figure}

\subsection{B. Supplementary Implementation Details}
\label{app:details}

\subsubsection{B.1 Details of Backbone Model}
\paragraph{Model and Hardware.}
All experiments are conducted with a frozen \textbf{Llama-3.1-8B-Instruct} language model (Huggingface Transformers), without any parameter updates or further pre-training. All inference was performed on a single NVIDIA Quadro RTX 8000 (48GB GPU), with the efficiency and resource analysis experiments in \textbf{Appendix D.4} additionally introducing an NVIDIA A100 (80GB GPU) for comparison.

\paragraph{Model Scale Considerations.}
We used a single backbone because, under AD data scarcity, the pre-training exposure is limited across vary scales. Nevertheless, model capacity is expected to modulate both absolute performance and the \emph{marginal} benefit of our components. 
We argue the following hypothesis:
\begin{itemize}
    \item Larger LLMs, endowed with stronger task perception and richer pre-trained priors, should deliver higher baselines; their \emph{absolute} gains from DCR and PVA will likely shrink due to diminishing returns, yet \emph{relative} gains can persist because both modules operate on high-quality hidden states (HS) and mainly improve decision calibration and context routing rather than raw capability.
    \item Smaller LLMs, with weaker HS quality, are expected to see larger \emph{absolute} gains, DCR widens salient cues in the prompt space and PVA deepens integration at separator tokens, while their upper bounds remain constrained by HS fidelity. 
    \item Across scales, we anticipate a monotone trade-off between compute/latency and robustness: deeper/multi-layer anchoring yields consistent improvements, but its cost is amortized more effectively on mid-to-large models that already maintain stable HS geometry. 
\end{itemize}
A full multi-scale analysis (e.g., 3B/8B/14B/70B) is left as future work with both absolute and relative gains reported.

\subsubsection{B.2 Details of Demo Construction}
\paragraph{Prompt Construction.}
All prompts are in English, following the standard in-context learning format (Details in Tab.~\ref{tab:prompt_template}) with demonstration pairs. Each prompt is constructed as a sequence of $k$ demonstration pairs, each formatted as ``Input~$\to$~Output'', followed by the test input and separator:
\begin{equation*}
[x_1~\to~y_1;\;x_2~\to~y_2;\;\ldots;\;x_k~\to~y_k;\;x_\mathrm{test}~\to~]
\end{equation*}
\vspace{-1ex}
A concrete illustration is shown below.

\paragraph{Demo Retrieval and Construction.}
Semantic similarity is computed using the cosine similarity between the test sample and candidate demos, based on the last hidden state of the model at the final token. Length similarity is measured by token count. Eight main demonstrations (4 AD / 4 Control, see \textbf{Appendix C}) are selected for each test sample via four orthogonal criteria (semantic/length, similarity/dissimilarity), with each main demo paired to a sub-demo for task vector extraction. Label balance is strictly enforced; no overlap occurs between main and sub-demo sets.

\paragraph{Hyperparameters and Inference.}
As illustrated in the PCA analysis (Fig.~\ref{fig:layerwise_pca}), middle layers undergo a rapid shift along PC-3 that then stabilizes near deep layers. Strong anchoring here can perturb downstream feature composition. Thus, the layer-wise injection strength $\gamma^{(l)}$ is set to 1.0 for the first and last 8 layers, and 0.2 for the middle 16 layers. Each experiment is repeated for 10 runs with different random seeds. All other hyper-parameters are listed in Tab.~\ref{tab:hyperparam}.

\subsubsection{B.3 Details of Experimental settings}
\paragraph{Statistical Analysis.}
Results are reported as mean $\pm$ standard deviation over 10 independent runs. Paired t-tests and Wilcoxon signed-rank tests are computed via \texttt{scipy.stats}. No multiple-comparison correction is applied, since only one-to-one comparisons are made per setting.

\paragraph{Evaluation Metrics.}
All results are reported on metrics (Accuracy, Precision, Recall, F1). For each run, metrics are computed on the test set and then averaged.

\begin{figure}[!ht]
\centering
\includegraphics[width=\linewidth]{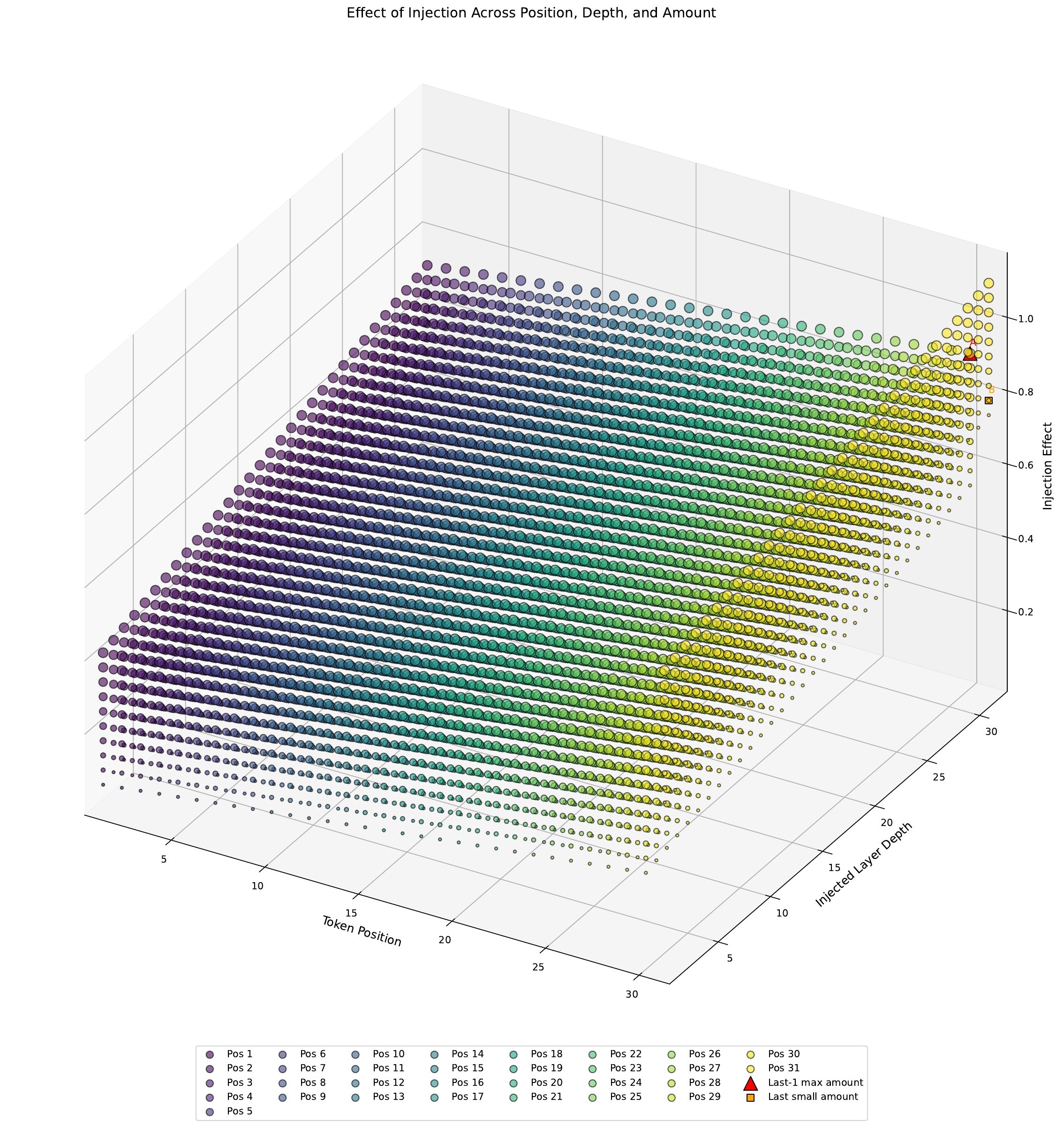}
\caption{Conceptual illustration of Task Vector (TV) injection effects. 
The x-axis denotes the token position, the y-axis indicates injected layer depth, and the z-axis (color) represents the relative injection effect. This figure serves as a schematic visualization to clarify the propagation pattern rather than an empirical result. Injections applied to the final (anchor) token accumulate and dominate downstream influence, while earlier token injections have weaker or dispersed effects.}
\label{fig:tv_injection_effects}
\end{figure}

\begin{table}[ht]
\centering
\begin{tabular}{l|c}
\toprule
Parameter & Value \\
\midrule
Backbone Model & Llama-3.1-8B-Instruct (frozen) \\
Max sequence length & 512 \\
Demo count per test & 8 (4 pairs) \\
Batch size & 1 \\
Layer-wise $\gamma$ & 1.0 (0-7)/ 0.2 (8-23)/ 1.0 (24-31) \\
Runs per experiment & 10 \\
GPU & Quadro RTX 8000, 48GB \\
Metrics & Precision, Recall, F1, Accuracy \\
\bottomrule
\end{tabular}
\caption{Main hyperparameters and computational resources.}
\label{tab:hyperparam}
\end{table}

\subsection{C. Supplementary Mechanistic Analysis}
\label{app:mechanism}

To rigorously complement the theoretical foundation discussed in the technical content of our paper, this section provides a detailed and formalized analysis of the underlying mechanisms of In-Context Learning (ICL) and Task Vector (TV) methods from the perspective of the Transformer’s residual stream formulation, alongside an extended discussion on TV injection effects.

\subsubsection{C.1 In-Context Learning (ICL): Residual Stream Formulation}

\paragraph{Residual Stream in Transformers.} 
Following previous formulations, we represent the hidden state at token $t$ and layer $l$ as part of a recursive residual stream:
\begin{equation}
h_t^{(l)} = h_t^{(l-1)} + a_t^{(l)} + m_t^{(l)},
\end{equation}
where $a_t^{(l)}$ and $m_t^{(l)}$ denote multi-head attention (MHA) and multi-layer perceptron (MLP) modules, respectively. The attention mechanism aggregates information across tokens:
\begin{equation}
\small
\begin{aligned}
a_t^{(l)} &= \mathrm{MHA}(h_t^{(l-1)})\\
&= \sum_{i=1}^T \mathrm{softmax}\left(\frac{(W_Q h_t^{(l-1)})^\top (W_K h_i^{(l-1)})}{\sqrt{d}}\right) W_V h_i^{(l-1)},
\end{aligned}
\label{eqMHA}
\end{equation}
where $T$ is the sequence length and $W_Q, W_K, W_V$ are learnable projection matrices. The MLP module introduces non-linear transformations:
\begin{equation}
m_t^{(l)} = W_2\left(\sigma(W_1(h_t^{(l-1)} + a_t^{(l)}))\right),
\end{equation}
where $\sigma$ is a nonlinear activation function. Collectively, these mechanisms progressively refine hidden representations layer-by-layer.

\paragraph{Generation Flow in ICL.}
In ICL, given a demonstration set $D = \{(x_i,y_i)\}_{i=1}^k$ concatenated with a test query $x_{\text{test}}$:
\begin{equation}
\mathrm{Seq}(D) = [x_1\!\to y_1;\dots;x_k\!\to y_k; x_{\text{test}}\!\to],
\end{equation}
the model predicts labels sequentially, with the next-token probability:
\begin{equation}
p(x_t \mid x_{<t}) = \mathrm{softmax}(W_{LM} h_{t-1}^{(L)}),
\end{equation}
where $W_{LM}$ is the final projection matrix mapping hidden states to token logits.

\paragraph{Final Token as Contextual Anchor.}
Although each separator token ($\to$) locally integrates contextual information within individual demos, the final token at position $T$ in the last layer ($L$) serves as the \textbf{global anchor}:
\begin{equation}
h_{\to, T}^{(L)} = \mathrm{TransformerLayer}^{(L)}([h_1^{(L-1)}, \dots, h_{T}^{(L-1)}]).
\end{equation}
This global anchor token integrates and consolidates all previously aggregated context across demonstrations and the test input, thus critically influencing model predictions.

\paragraph{Layer-wise Information Refinement.}
The Transformer progressively transforms representations from general and shallow features in early layers to increasingly specialized and discriminative task-related representations in deeper layers. This hierarchical refinement ensures the final-layer representations are rich in task-specific semantics.

\subsubsection{C.2 Task Vector (TV) Method: Extraction and Injection Mechanism}

\paragraph{Task Vector Extraction.}
Standard TV methods extract task-specific latent vectors at a chosen intermediate layer $\ell$ by concatenating a demo set with a randomly chosen pseudo-query $x_{\mathrm{pseudo}}$:
\begin{equation}
\mathrm{Seq}(D) = [x_1\!\to y_1;\dots;x_k\!\to y_k;x_{\mathrm{pseudo}}\!\to].
\end{equation}
The extracted task vector $v^{(\ell)}$ is then:
\begin{equation}
v^{(\ell)} = h_{\to,\text{pseudo}}^{(\ell)},
\end{equation}
where $h_{\to,\text{pseudo}}^{(\ell)}$ represents the hidden state at the pseudo-query separator token.

\paragraph{Single-point Task Vector Injection.}
During inference, the task vector is injected into the separator token of the test input at layer $\ell$:
\begin{equation}
h_{\to,\text{test}}^{(\ell)\prime} = f(h_{\to,\text{test}}^{(\ell)}, v^{(\ell)}),
\end{equation}
where $f(\cdot)$ can be soft (additive, $0 \leq \gamma < 1$) or hard (replacement, $\gamma = 1$):
\begin{equation}
\small
f(h, v) = (1 - \gamma) \cdot h + \gamma \cdot v.
\end{equation}

\paragraph{Vertical Propagation and Horizontal Limitation.}
The injected vector modifies hidden states and propagates vertically across subsequent layers:
\begin{equation}
h_{\to,\text{test}}^{(\ell+1)} = \mathrm{TransformerLayer}^{(\ell+1)}([h_1^{(\ell)}, \dots, h_{\to,\text{test}}^{(\ell)\prime}]).
\end{equation}
However, horizontally, the injected vector only influences subsequent layers at the test position, significantly limiting contextual integration capabilities and potentially causing pseudo-sample overfitting, especially when using single or shallow-layer injection.

\begin{table*}[!ht]
\centering
\begin{tabular}{l|c|cc|cc|cc}
\toprule
    \multirow{2}{*}{Methods} 
  & \multirow{2}{*}{Sig.}
  & \multicolumn{2}{c|}{Test} 
  & \multicolumn{2}{c|}{Lu} 
  & \multicolumn{2}{c}{Pitt} \\
  &   & F1 (\%) & Acc (\%) & F1 (\%) & Acc (\%) & F1 (\%) & Acc (\%) \\
\midrule
ICL$_{Van}$   & --    & 72.13$_{4.15}$ & 70.00$_{4.08}$ & 79.53$_{3.58}$ & 70.95$_{5.08}$ & 70.24$_{1.12}$ & 66.48$_{1.20}$  \\
+Van PVA    & \ding{55}      & 73.75$_{4.77}$ & 70.83$_{5.19}$ & 78.01$_{2.44}$ & 67.62$_{3.87}$ & 70.32$_{1.43}$ & 66.19$_{1.84}$  \\
+Sem PVA    & \ding{55}      & 72.72$_{6.43}$ & 70.21$_{6.46}$ & 79.09$_{2.53}$ & 69.52$_{2.97}$ & 71.68$_{1.26}$ & 67.54$_{1.43}$  \\
+DCR PVA    & \ding{55}      & 70.17$_{2.94}$ & 66.88$_{2.18}$ & 80.54$_{3.05}$ & 71.90$_{4.49}$ & 71.54$_{1.40}$ & 67.49$_{1.61}$  \\
\midrule
ICL$_{Sem}$   & --    & 74.20$_{3.69}$ & 71.04$_{4.32}$ & 82.02$_{2.11}$ & 73.33$_{3.50}$ & 73.79$_{1.21}$ & 69.71$_{1.13}$  \\
+Van PVA     & \ding{55}     & 75.76$_{3.61}$ & 72.71$_{4.00}$ & 80.06$_{0.96}$ & 69.76$_{1.52}$ & 73.91$_{1.08}$ & 69.73$_{1.41}$  \\
+Sem PVA     & \ding{55}      & 77.20$_{3.91}$ & 74.58$_{4.82}$ & 79.81$_{2.75}$ & 69.76$_{4.13}$ & 73.03$_{0.46}$ & 68.96$_{0.48}$ \\
+DCR PVA    & \ding{55}      & 77.20$_{3.30}$ & 74.79$_{3.66}$ & 81.16$_{1.40}$ & 71.67$_{2.49}$ & 73.90$_{0.99}$ & 69.89$_{1.26}$ \\
\midrule
ICL$_{DCR}$   & --    & 76.75$_{4.53}$ & 77.29$_{4.41}$ & 83.71$_{2.11}$ & 77.86$_{2.83}$ & 76.62$_{1.41}$ & 75.94$_{1.21}$  \\
+Van PVA    & \ding{51}      & 80.78$_{5.05}$ & 80.62$_{5.27}$ & 84.82$_{2.34}$ & 79.05$_{3.16}$ & 79.86$_{1.13}$ & 78.96$_{1.09}$  \\
+Sem PVA     & \ding{51}      & 80.56$_{4.11}$ & 80.42$_{4.08}$ & 85.41$_{1.78}$ & 80.00$_{2.43}$ & 80.16$_{0.59}$ & 79.22$_{0.62}$ \\
+DCR PVA (Ours) & \ding{51}  & 86.11$_{1.92}$ & 85.83$_{1.91}$ & 86.51$_{1.83}$ & 81.43$_{2.56}$ & 80.23$_{0.42}$ & 79.42$_{0.39}$ \\
\bottomrule
\end{tabular}
\caption{Ablation results for the PVA module under different demo retrieval settings (4 demo pairs) on Test, Lu, and Pitt datasets. Each cell reports mean$_{\text{std}}$ over 10 runs. “Sig” indicates whether the PVA variant significantly outperforms its base ICL (\ding{51}) or not (\ding{55}).}
\label{tab:pva_ablation_full}
\end{table*}

\subsubsection{C.3 Visualizing Limitations of TV Injection}

To clarify the impact of injection design, we visually illustrate how TV effectiveness is influenced by three key design choices: injection position, injection depth, and the number of injection layers (Fig.~\ref{fig:tv_injection_effects}). In this illustration:

\begin{itemize}
    \item \textbf{Position (x-axis):} Indicates the relative token position of injection (left: early, right: late).
    \item \textbf{Layer Depth (y-axis):} Denotes injection at shallow (front) versus deep (behind) layers.
    \item \textbf{Injection Breadth (Sphere size):} Larger spheres represent injection spanning more layers; smaller spheres reflect fewer layers (or single-layer injection).
\end{itemize}

This visualization clearly demonstrates that:
\begin{itemize}
    \item Injecting at very shallow layers has limited vertical propagation potential, even high quality of TV, offering insufficient task specialization.
    \item A bad Injection at deep layers may disrupt context-sensitive representations, causing semantic distortion or overfitting.
    \item Single-layer or limited-layer injection results in restricted horizontal propagation, failing to enrich broader contextual understanding adequately.
\end{itemize}
Thus, the TV method's effectiveness critically depends on balanced choices regarding these dimensions. The limitations inherent in conventional single-layer, test-centric injection motivate our proposed demo-centric, multi-layer projection anchoring strategy, which distributes and harmonizes task adaptation signals more effectively across both horizontal and vertical dimensions.

This rigorous preliminary formalization and expanded visualization collectively provide clear theoretical grounding for our proposed methodological improvements and elucidate the necessity of adopting demo-centric, multi-layer injection approaches for robust adaptation in nuanced tasks such as Alzheimer's disease detection.

\begin{figure}[ht]
\centering
\includegraphics[width=\columnwidth]{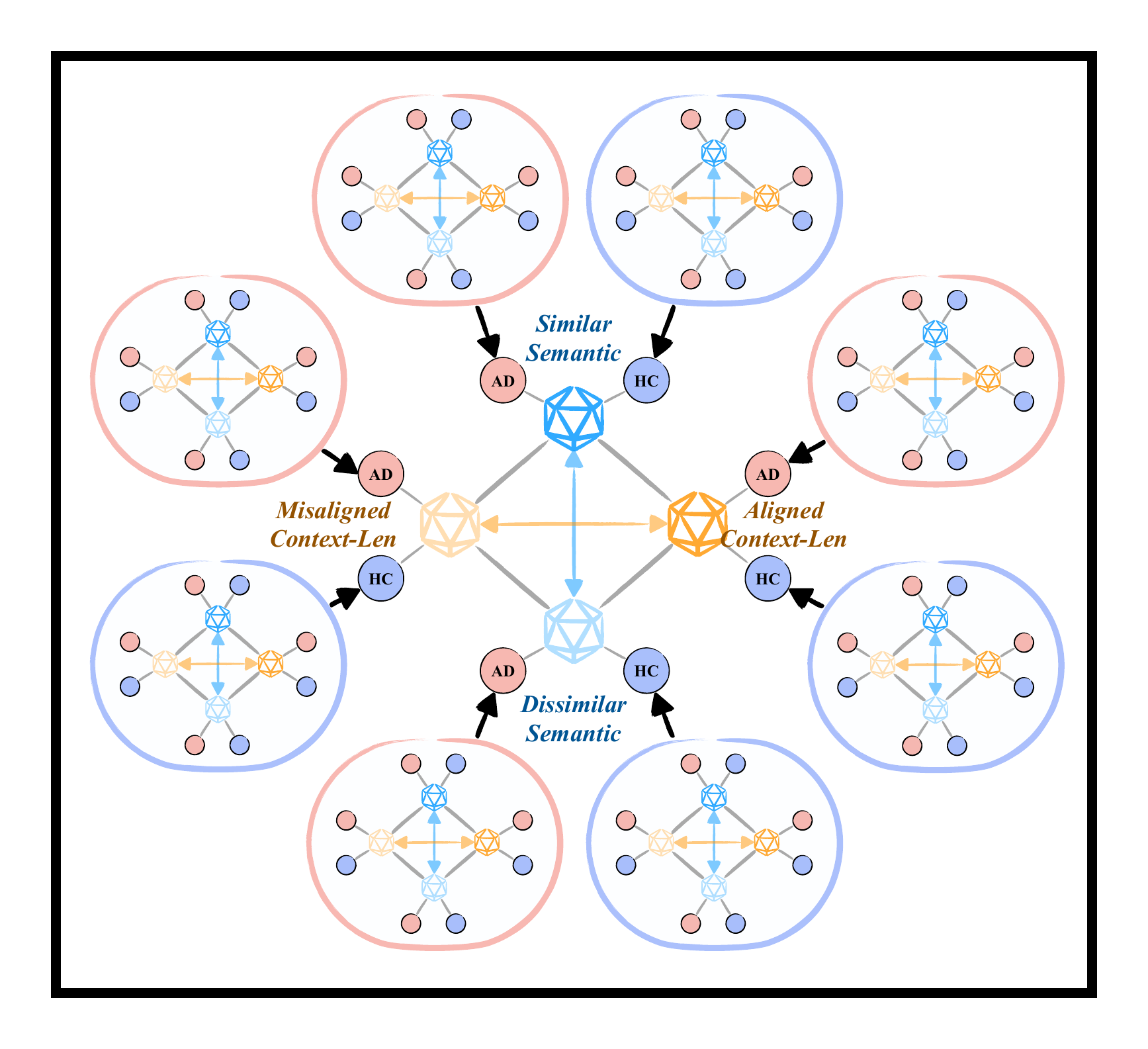}
\caption{Demo-centric, projection-based full-layer anchoring: Each main demo separator is injected with its specific task vector at all layers, enabling distributed and context-rich adaptation.}
\label{fig:concept_format}
\end{figure}

\subsubsection{C.4 Demo Construction: Diverse and Contrastive Retrieval}

A core component of our approach is the construction of a high-diversity, contrastively balanced demo set for each test query via a two-stage retrieval mechanism. 

\paragraph{Stage 1: Main-Demo Set Construction.} 
For every test sample $d_{\mathrm{test}}$, we retrieve $8$ main demos, sampling one contrastive pair (AD/HC) for each of four orthogonal criteria:

\begin{itemize}
  \item \textbf{Semantic Similarity:} Pair with the highest cosine similarity in $\phi$-space (where $\phi$ is an LLM-derived embedding).
  \begin{equation}
    \begin{aligned}
      d^{\textit{sim}}_{+}
      &= \mathop{\arg\max}_{(x,y)\in\mathcal{S},y=AD}
      \cos\bigl(\phi(d_{\mathrm{test}}),\,\phi((x,y))\bigr),  \\
      d^{\textit{sim}}_{-}
      &= \mathop{\arg\max}_{(x,y)\in\mathcal{S},y=HC}
      \cos\bigl(\phi(d_{\mathrm{test}}),\,\phi((x,y))\bigr)
    \end{aligned}
  \end{equation}
  \item \textbf{Semantic Dissimilarity:} Pair with the lowest cosine similarity.
  \begin{equation}
    \begin{aligned}
      d^{dis}_{+}
      &= \mathop{\arg\min}_{(x,y)\in\mathcal{S},y=AD}
      \cos\bigl(\phi(d_{\mathrm{test}}),\,\phi((x,y))\bigr), \\
      d^{dis}_{-}
      &= \mathop{\arg\min}_{(x,y)\in\mathcal{S},y=HC}
      \cos\bigl(\phi(d_{\mathrm{test}}),\,\phi((x,y))\bigr)
    \end{aligned}
  \end{equation}
  \item \textbf{Length Similarity:} Pair whose text length is closest to $d_{\mathrm{test}}$.
  \begin{equation}
    \begin{aligned}
      d^{\textit{sim}\ell}_{+}
      &= \mathop{\arg\min}_{(x,y)\in\mathcal{S},y=AD}
      \bigl|\ell(d_{\mathrm{test}})-\ell((x,y))\bigr|,\\
      d^{\textit{sim}\ell}_{-}
      &= \mathop{\arg\min}_{(x,y)\in\mathcal{S},y=HC}
      \bigl|\ell(d_{\mathrm{test}})-\ell((x,y))\bigr|
    \end{aligned}
  \end{equation}
  \item \textbf{Length Dissimilarity:} Pair whose length is most distant.
  \begin{equation}
    \begin{aligned}
      d^{\textit{dis}\ell}_{+}
      &= \mathop{\arg\max}_{(x,y)\in\mathcal{S},y=AD}
      \bigl|\ell(d_{\mathrm{test}})-\ell((x,y))\bigr|,\\
      d^{\textit{dis}\ell}_{-}
      &= \mathop{\arg\max}_{(x,y)\in\mathcal{S},y=HC}
      \bigl|\ell(d_{\mathrm{test}})-\ell((x,y))\bigr|
    \end{aligned}
  \end{equation}
\end{itemize}
Here, $\ell(d)$ denotes the sequence length of demo $d$. 
\noindent
Together, the selected demos form the main demo set:
\begin{equation}
    \mathcal{D}_{\mathrm{main}}(d_{test}) = \{d^{\textit{sim}}_{+}, d^{\textit{sim}}_{-}, d^{dis}_{+}, d^{dis}_{-}, d^{\textit{sim}\ell}_{+}, d^{\textit{sim}\ell}_{-}, d^{\textit{dis}\ell}_{+}, d^{\textit{dis}\ell}_{-}\}.
\end{equation}

\paragraph{Stage 2: Sub-Demo Set Construction.}
For each main demo $d_i \in \mathcal{D}_{\mathrm{main}}$ from Stage~1, we construct a \emph{sub-demo set} to extract a demo-centric TV. The retrieval criterion is
same with Stage~1. Concretely, for a main demo $d_{\text{main}}$, we select:
\begin{equation}
   \mathcal{D}_{\mathrm{sub}}(d_{i}) = \{d^{\textit{sim}}_{+}, d^{\textit{sim}}_{-}, d^{dis}_{+}, d^{dis}_{-}, d^{\textit{sim}\ell}_{+}, d^{\textit{sim}\ell}_{-}, d^{\textit{dis}\ell}_{+}, d^{\textit{dis}\ell}_{-}\}.
\end{equation}

As Fig.~\ref{fig:concept_format} shows, repeated for each dimension and balanced across AD/HC labels, yields a diverse set of 8 demos per test sample, providing a rich, contrastive context by covering the semantic and structural extremes of the demo space. All similarities are computed directly using the final-layer hidden representations of the LLM, ensuring alignment with the model’s own reasoning space and avoiding external embedding mismatch. This design enables the prompt to cover both “close” and “distant” neighbors in both semantic and structural spaces, ensuring that the few-shot context exposes the model to a spectrum of possible features, mitigating overfitting and improving generalization.

\begin{table}[!ht]
\centering
\setlength{\tabcolsep}{13pt}
\begin{tabular}{l|c}
\toprule
\textbf{Hyperparameters} & \textbf{Value} \\
\midrule
\textbf{Maximum Input Length} & 1024 tokens \\
\textbf{Maximum New Tokens} & 105 tokens \\
\textbf{Temperature} & 0.95 \\
\textbf{Top-p Sampling} & 0.7 \\
\textbf{Batch Size} & 2 \\
\bottomrule
\end{tabular}
\caption{Hyperparameters for LoRA fine-tuning and evaluation.}
\label{tab:lora_hyperparameters}
\end{table}

\begin{table*}[!ht]
\centering
\setlength{\tabcolsep}{4pt}
\begin{tabular}{l c | cccc | cccc | cccc}
\toprule
\multirow{2}{*}{\textbf{Method}} & \multirow{2}{*}{\textbf{Shot}} &
\multicolumn{4}{c|}{\textbf{Test} (\%)} &
\multicolumn{4}{c|}{\textbf{Lu} (\%)} &
\multicolumn{4}{c}{\textbf{Pitt} (\%)} \\
 & & Acc & Pre & Rec & F1 & Acc & Pre & Rec & F1 & Acc & Pre & Rec & F1 \\
\midrule
\textbf{BERT$_c$}      & N/A & 81.25 & \underline{94.11} & 66.67 & 78.05 & \textbf{83.33} & \textbf{85.71} & 88.89 & \underline{87.27} & 70.12 & 82.87 & 58.49 & 68.58 \\
\textbf{BERT$_f$}      & N/A & 81.25 & 82.61 & 79.17 & 80.85 & \underline{80.95} & 80.64 & \underline{92.59} & 86.21 & 77.05 & 85.16 & 71.24 & 77.58 \\
\textbf{SPZ}           & N/A & \textbf{91.66} & \textbf{95.45} & 87.50 & \textbf{91.30} & \textbf{83.33} & \underline{83.33} & \underline{92.59} & \textbf{87.72} & \underline{77.96} & \textbf{88.07} & 69.93 & \underline{77.96} \\
\midrule
\textbf{LoRA$_{cls}$}  & N/A & 60.40 & 55.81 & \textbf{100.00} & 71.64 & 64.29 & 65.00 & \textbf{96.29} & 77.61 & 57.92 & 57.19 & \textbf{97.39} & 72.07 \\
\textbf{LoRA$_{gen}$}  & N/A & 41.67 & 43.33 & 54.17 & 48.15 & 47.62 & 60.87 & 51.85 & 56.00 & 51.18 & 57.19 & 49.35 & 52.98 \\
\midrule
\textbf{DA4ICL (Ours, best)} & 8 & \underline{89.58} & 88.00 & \underline{91.67} & \underline{89.80} 
                                & \textbf{83.33} & \underline{83.33} & \underline{92.59} & 87.22 
                                & \textbf{80.33} & \underline{86.67} & \underline{76.47} & \textbf{81.25} \\
\bottomrule
\end{tabular}
\caption{Representative supervised baselines on Test / Lu / Pitt, and DA4ICL result. Bold entries denote the best result, and underlined entries denote the second-best. }
\label{tab:supervised_bibm_combined}
\end{table*}

\subsection{D. Supplementary Experiments}

\subsubsection{D.1 Ablation Results for PVA Module}

Tab.~\ref{tab:pva_ablation_full} presents the ablation study for the Projected Vector Anchoring (PVA) module under various demo retrieval settings (4 demo pairs) on Test, Lu, and Pitt datasets. Statistical significance is indicated for each variant.

\subsubsection{D.2 Full Evaluation Metrics}

Tab.~\ref{tab:main_results_full} reports the complete evaluation results (Precision, Recall, F1-score, and Accuracy) for all methods across the Test, Lu, and Pitt datasets, with varying numbers of demonstrations ($N$). Each entry shows the mean and standard deviation over multiple runs.

\subsubsection{D.3 Comparison of Supervised Baselines}
As an additional comparison, Tab.~\ref{tab:supervised_bibm_combined} reports representative supervised baselines alongside the DA4ICL result. All reported results are based on the best-performing run.

\paragraph{Fine-tuning Methods}
1) BERT (coarse-grained) \cite{balagopalan2021bert}: Utilizes the CLS token for classification;
2) BERT (fine-grained) \cite{balagopalan2021bert}: Extracts representations via GlobalMaxPooling;
3) SPZ \cite{spz}: Employs semantic perturbations and zonal-mixing to enhance robustness;
4) LoRA (CLS) \cite{hu2022lora}: Applies low-rank adaptation for parameter-efficient fine-tuning on LLM under classification task;
5) LoRA (GEN) \cite{hu2022lora}: Applies low-rank adaptation for parameter-efficient fine-tuning on LLM under text generation task.

\begin{table*}[!ht]
\centering
\begin{tabular}{l l c c c}
\toprule
\textbf{Setting} & \textbf{Variant} & \textbf{Elapsed (s)} & \textbf{Peak Alloc (MB)} & \textbf{Peak Reserved (MB)} \\
\midrule
\multicolumn{5}{l}{\textit{48GB GPU}} \\
\midrule
Vanilla ICL  & --                & $15.65 \pm 0.27$ & $16160.05 \pm 0.54$  & $16364.80 \pm 146.40$ \\
Semantic ICL & --                & $15.72 \pm 0.18$ & $16160.05 \pm 0.54$  & $16364.80 \pm 146.40$ \\
DCR ICL      & --                & $44.99 \pm 1.95$ & $16677.95 \pm 0.17$  & $16904.60 \pm 16.20$  \\
\midrule
Vanilla TV   & Add        & $60.47 \pm 1.24$ & $18618.89 \pm 225.60$ & $21126.00 \pm 997.70$ \\
             & Replace     & $63.17 \pm 5.18$ & $18567.99 \pm 134.72$ & $21090.80 \pm 1385.58$ \\
Semantic TV  & Add         & $79.13 \pm 1.75$ & $18630.14 \pm 141.89$ & $22666.00 \pm 942.08$ \\
             & Replace     & $71.29 \pm 4.92$ & $18594.08 \pm 120.34$ & $22926.80 \pm 1648.29$ \\
DCR TV       & Add         & $80.24 \pm 2.17$ & $19370.89 \pm 118.89$ & $21107.60 \pm 640.76$ \\
             & Replace     & $70.08 \pm 7.29$ & $19304.88 \pm 67.69$  & $21458.20 \pm 971.24$ \\
\midrule
DA4ICL       & --   & $521.63 \pm 42.68$ & $27865.15 \pm 0.00$ & $31651.80 \pm 17.40$ \\
\bottomrule
\end{tabular}
\caption{Runtime and peak GPU memory on a 48GB GPU. Mean $\pm$ SD over 10 runs.}
\label{tab:efficiency_48g}
\end{table*}

\begin{table*}[!ht]
\centering
\begin{tabular}{l l c c c}
\toprule
\textbf{Setting} & \textbf{Variant} & \textbf{Elapsed (s)} & \textbf{Peak Alloc (MB)} & \textbf{Peak Reserved (MB)} \\
\midrule
\multicolumn{5}{l}{\textit{80GB GPU}} \\
\midrule
Vanilla ICL  & --                & $12.64 \pm 2.17$ & $16501.42 \pm 29.30$  & $17053.00 \pm 226.22$ \\
Semantic ICL & --                & $20.36 \pm 0.74$ & $16545.22 \pm 16.65$  & $17033.80 \pm 135.46$ \\
DCR ICL      & --                & $26.35 \pm 0.35$ & $16678.37 \pm 1.16$   & $16971.60 \pm 10.80$  \\
\midrule
Vanilla TV   & Add         & $35.64 \pm 0.94$ & $17477.90 \pm 105.37$ & $19094.40 \pm 556.18$ \\
             & Replace     & $35.40 \pm 0.58$ & $17428.73 \pm 94.81$  & $19142.20 \pm 645.57$ \\
Semantic TV  & Add         & $36.65 \pm 0.96$ & $17442.27 \pm 38.88$  & $19734.00 \pm 492.66$ \\
             & Replace     & $37.14 \pm 1.10$ & $17463.38 \pm 61.94$  & $19531.60 \pm 532.08$ \\
DCR TV       & Add        & $41.20 \pm 0.61$ & $17897.43 \pm 105.93$ & $19091.20 \pm 490.37$ \\
             & Replace    & $41.33 \pm 0.83$ & $17866.03 \pm 52.96$  & $18852.60 \pm 329.37$ \\
\midrule
DA4ICL       & -- & $283.72 \pm 2.18$ & $26299.99 \pm 2.81$  & $27813.00 \pm 279.34$ \\
\bottomrule
\end{tabular}
\caption{Runtime and peak GPU memory on an 80GB GPU. Mean $\pm$ SD over 10 runs.}
\label{tab:efficiency_80g}
\end{table*}

\paragraph{Detailed Information of LoRA} \label{appendix:lora_details}
This section provides details of LoRA method used in our experiments. LoRA is evaluated under two supervised paradigms: \textbf{classification tasks} and \textbf{generation tasks}, to assess its adaptability in AD detection scenarios. Hyperparameters and task-specific configurations are detailed below.

\textbf{Hyperparameters.}
For hyperparameters like \textit{lora-rank ($\gamma$)}, \textit{lora-alpha ($\alpha$)} and \textit{target modules}\footnote{Here we set $\gamma=16$, $\alpha=8$ and set all attention modules (Q,K,V,O) as target modules.}, we use default settings to ensure consistency and robustness across experiments. Tab.~\ref{tab:lora_hyperparameters} summarizes the hyperparameters used for fine-tuning and evaluating LoRA in both task paradigms.

\textbf{LoRA Classification Task Setup.}
In the classification task, LoRA is trained to differentiate between AD and normal cognitive states based on textual descriptions of the "Cookie Theft" picture. Each data instance comprises an instruction, an input description, and an output label. The details of prompt construction in classification task are shown in Tab.~\ref{tab:lora_classification}.

\textbf{LoRA Generation Task Setup.}
In the generation task, LoRA is supervised to produce textual descriptions based on specified mental states (e.g., alzheimer's disease or normal cognitive conditions). This paradigm evaluates the model's capacity to simulate descriptive outputs conditioned on cognitive states. The details of prompt construction in classification task are shown in Tab.~\ref{tab:lora_generation}.

The classification-based LoRA evaluates the LLMs' ability to differentiate between AD and control transcripts by assigning discrete labels, while the generation-based LoRA assesses its capacity to simulate descriptive patterns under specified cognitive conditions. These complementary configurations ensure a robust evaluation of LoRA’s adaptability and effectiveness across diverse tasks relevant to AD detection.

\paragraph{Analysis of Results}

\textbf{Comparison with supervised SOTA.}
While our scope targets LLM adaptation under limited task cognition and weak contextual perception (hence the emphasis on ICL/TV families), we additionally benchmark against representative supervised baselines (Tab.~\ref{tab:supervised_bibm_combined}) to address practical utility. On \textit{Test} dataset, DA4ICL attains second-best Accuracy/F1, trailing only SPZ, and on \textit{Lu} corpus, it matches the best Accuracy and remains near-best F1. On the large OOD \textit{Pitt} corpus, it achieves the top Accuracy and F1. These results indicate that, despite using a frozen backbone with only 8-shot prompting (no gradient updates), DA4ICL is competitive with fine-tuned encoders (BERT variants) and strong data augmentation methods (SPZ), and surpasses LoRA fine-tuning settings under both classification and generation paradigms. In short, DA4ICL narrows the gap to supervised SOTA on ID-style splits and provides stronger transfer on OOD corpora, a property that is critical for real clinical deployment where domain shift is the norm.

\textbf{Why this matters for medical use.}
Supervised pipelines typically require labeled training, multiple training runs, and careful regularization to avoid overfitting to specific corpus. DA4ICL instead operates in a label-efficient, training-free alternative, where DCR widens salient cues in the prompt space while PVA deepens multi-layer context integration at anchor positions, yielding better cross-corpus calibration without additional training data or parameters updates. Practically, this reduces development friction (e.g., no re-training when moving across clinics/corpora) and lowers the risk of distributional mismatch.

\textbf{Complementary strengths.}
Encoder-based supervised baselines (e.g., BERT$_c$/BERT$_f$) excel when train/test share distribution, leveraging discriminative representations learned from labels; DA4ICL’s gains stem from improved \emph{context routing} inside a frozen LLM, which is inherently robust to label scarcity and lexical drift. The patterns in Tab.~\ref{tab:supervised_bibm_combined} reflect this complementarity: SPZ leads on smaller, cleaner splits (\textit{Test}), while DA4ICL dominates on \textit{Pitt} where stylistic and demographic variability amplify OOD shift.

\begin{figure}[!ht]
\centering
\includegraphics[width=\linewidth]{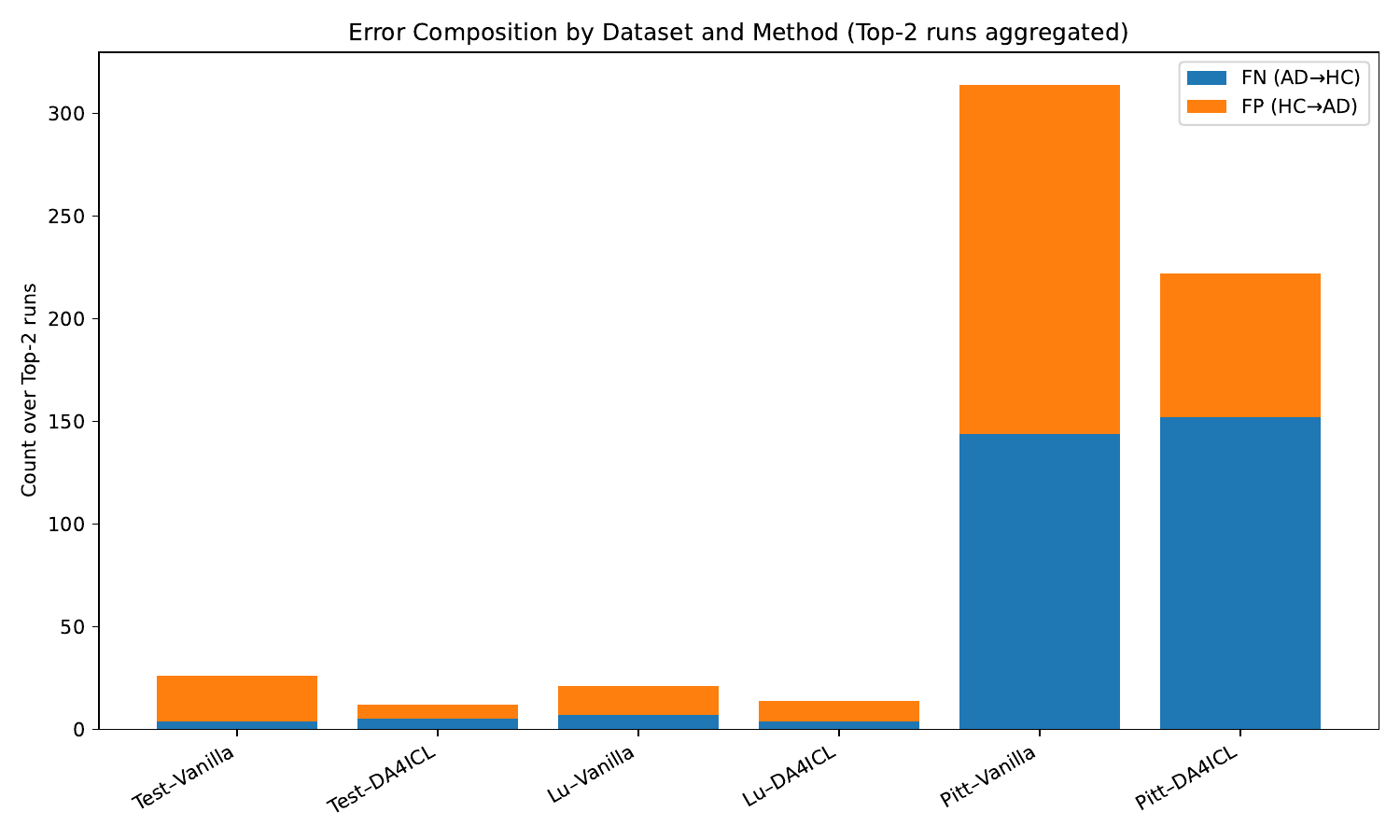}
\caption{Error composition across datasets for the Top-2 performing runs of Vanilla ICL and DA4ICL. 
Each bar represents the sum of false negatives (FN, AD$\!\to$HC) and false positives (FP, HC$\!\to$AD) across Top-2 performing runs. DA4ICL exhibits a clear reduction in total errors and a more symmetric FN–FP distribution across all datasets, indicating improved contextual calibration and robustness, particularly under out-of-distribution conditions (Lu and Pitt).}
\label{fig:error_analysis}
\end{figure}

\begin{table*}[!ht]
\small
\centering
\begin{tabular}{l l c c c c c c c c}
\toprule
\textbf{Dataset} & \textbf{Method} & \textbf{Run} & \textbf{N} & \textbf{Errors} & \textbf{FN (1$\to$0)} & \textbf{FP (0$\to$1)} & \textbf{Errors (\%)} & \textbf{FN (\%)} & \textbf{FP (\%)} \\
\midrule
Test & ICL Van & 1 & 48 & 13 & 4 & 9 & 27.08 & 8.33 & 18.75 \\
Test & ICL Van & 2 & 48 & 13 & 0 & 13 & 27.08 & 0.00 & 27.08 \\
Test & DA4ICL  & 1 & 48 & 5  & 2 & 3  & 10.42 & 4.17 & 6.25 \\
Test & DA4ICL  & 2 & 48 & 7  & 3 & 4  & 14.58 & 6.25 & 8.33 \\
\midrule
Lu   & ICL Van & 1 & 32 & 9  & 2 & 7  & 28.12 & 6.25 & 21.88 \\
Lu   & ICL Van & 2 & 32 & 12 & 5 & 7  & 37.50 & 15.62 & 21.88 \\
Lu   & DA4ICL  & 1 & 32 & 7  & 2 & 5  & 21.88 & 6.25 & 15.62 \\
Lu   & DA4ICL  & 2 & 32 & 7  & 2 & 5  & 21.88 & 6.25 & 15.62 \\
\midrule
Pitt & ICL Van & 1 & 549 & 152 & 66 & 86 & 27.69 & 12.02 & 15.67 \\
Pitt & ICL Van & 2 & 549 & 162 & 78 & 84 & 29.60 & 14.21 & 15.30 \\
Pitt & DA4ICL  & 1 & 549 & 108 & 72 & 36 & 19.67 & 13.12 & 6.56 \\
Pitt & DA4ICL  & 2 & 549 & 114 & 80 & 34 & 20.76 & 14.57 & 6.19 \\
\bottomrule
\end{tabular}
\caption{Top-2 runs error details per dataset and method. Percentages are relative to N.}
\label{tab:top2_error_details}
\end{table*}

\textbf{Limitations and scope.}
Our supervised comparisons focus on widely used text-based baselines (BERT family, SPZ) and parameter-efficient LLM adaptation (LoRA), yet broader clinical SOTA also includes multi-modal systems and task-specific Transformers. A full audit of such systems, especially those using audio cues, falls outside our text-only scope but is a valuable direction. Integrating paralinguistics and multi-modal descriptors into the DCR\&PVA framework and expand cross-corpus validation to additional clinical cohorts to further probe generalization under real-world shift is promising in future works.

\subsubsection{D.4 Efficiency and Resource Analysis}
To assess computational overhead and scalability, we measured wall-clock runtime and peak GPU memory during inference for all baselines and our proposed DA4ICL under identical hardware conditions (48GB and 80GB NVIDIA GPUs). All measurements were obtained using \texttt{torch.cuda.max\_memory\_allocated()} and \texttt{max\_memory\_reserved()} after resetting CUDA memory statistics at the start of each run. 
As summarized in Tab.~\ref{tab:efficiency_48g} and Tab.~\ref{tab:efficiency_80g}, standard ICL variants (Vanilla, Semantic retrieval) exhibit lightweight computational overhead, requiring approximately 16--17\,GB of allocated memory and 15--26\,s of runtime. 
DCR retrieval introduces additional retrieval and prompt-assembly cost, slightly increasing inference time but maintaining comparable memory usage. 
Task Vector (TV) methods incur higher resource demand due to per-layer injection and fusion operations, leading to 18--19\,GB allocation and roughly double runtime relative to pure ICL. 
In contrast, the full DA4ICL pipeline, combining multi-layer projection, demo-centric anchoring, and retrieval, represents the most resource-intensive configuration (up to 27.9\,GB on 48GB GPU and 26.3\,GB on 80GB GPU), yet remains stable on a single GPU. 
These results confirm that DA4ICL achieves substantial performance gains with moderate and manageable computational overhead, preserving practical deployability in real-world low-resource inference settings.

\subsubsection{D.5 Error Analysis}
To further examine robustness and failure patterns, we analyzed the \textbf{Top-2 performing runs} of both Vanilla ICL and DA4ICL on the \textit{Test}, \textit{Lu}, and \textit{Pitt} datasets (See Fig.~\ref{fig:error_analysis} and Tab.~\ref{tab:top2_error_details}). Each dataset was evaluated by counting (i) total misclassifications and (ii) the direction of error: false negatives (AD$\!\to$HC) and false positives (HC$\!\to$AD). 

\paragraph{Overall Trends.}
DA4ICL consistently reduced total error counts across all datasets and runs. On the \textit{Test} set (48 samples), total errors dropped from 13/13 under Vanilla ICL to 5/7 under DA4ICL; on \textit{Lu} (32 samples), from 9/12 to 7/7; and on the large-scale \textit{Pitt} corpus (549 samples), from 152/162 to 108/114. This reduction demonstrates improved stability and generalization, especially under out-of-distribution (OOD) shifts.

\paragraph{Error Direction Asymmetry.}
Vanilla ICL exhibited a strong asymmetry in error direction, disproportionately over-predicting the positive (AD) class. For instance, in the \textit{Test} runs, false positives (HC$\!\to$AD) occurred 9/13 times versus only 4/0 false negatives, suggesting an over-reliance on pathological lexical cues. Similarly, on \textit{Pitt}, false positives (86/84) exceeded false negatives (66/78). In contrast, DA4ICL largely balanced this distribution: e.g., on \textit{Test}, errors were nearly symmetric (2–3 AD$\!\to$HC vs. 3–4 HC$\!\to$AD), and on \textit{Lu}, the ratios matched exactly (2–2 vs. 5–5). This balance indicates that the proposed demo-centric anchoring improves contextual calibration rather than merely shifting the decision boundary.

\paragraph{Qualitative findings.}
Beyond aggregate counts, the Top-2 misclassified transcripts reveal three recurring failure patterns that persist even under DA4ICL.
\begin{itemize}
    \item \textit{\textbf{Lexical stereotype bias}}: utterances richly mentioning canonical cues (overflowing sink, tipping stool, cookie jar) with otherwise fluent, orderly narration are sometimes predicted as HC even when their micro-structure shows AD-like markers (frequent fillers/repairs "uh/hm...", local incoherence, object substitutions such as \emph{ladder} for stool). This drives residual AD$\to$HC errors: when surface content units align strongly with the picture’s \textbf{gold schema}, the model’s decision can be dominated by content coverage rather than form.
    \item \textit{\textbf{Noisy idiosyncrasy as pathology}}: rare insertions or off-picture intrusions (e.g., joking or fantastical clauses like a “race horse jumping through the window”), plus pet phrases (“what the devil is~xxx?"), inflate HC$\to$AD false positives by mimicking semantic drift and pragmatic breakdown. DA4ICL reduces but does not eliminate this tendency.
    \item \textit{\textbf{Information density vs. sequencing}}: very concise, list-like descriptions that enumerate entities without linking actions/relations (weak event chaining, minimal causal connectives) remain ambiguous. When the final anchor token aggregates a short, weakly structured context, both ICL and DA4ICL can be brittle to small lexical fluctuations, yielding paired FP/FN flips across runs.
\end{itemize}

\begin{table*}[!ht]
\small
\centering
\begin{tabular}{lp{0.75\linewidth}}
\toprule
\textbf{Prompt Component} & \textbf{Details} \\
\midrule
System Prompt&
\texttt{You are an experienced clinician specializing in dementia care. You will analyze descriptions provided by subjects who are describing a scene from the "Cookie Theft" picture, used in the Boston Diagnostic Aphasia Examination. Your task is to analyze the subject's description to determine whether it indicates Alzheimer or Normal. Provide your answer in the format: \textbf{[A. Alzheimer Disease]} or \textbf{[B.Control]}.} \\
\midrule
Input & 
\texttt{Here is the example: [I see uh two kids up at the cookie jar, one on a stool the other standing on the floor. cupboard door is opened. mother's washing the dishes. the water is running overflowing the sink. and uh there's two cups and a plate on the counter. and she's washing holding a plate in her hand. curtains at the windows. the cookie jar has the lid off. hm hm that's about it. cupboards underneath the sink. cupboards underneath the other cupboards. uh kid falling off the stool. the girl laughing at him. cookies in the cookie jar with the lid off. he has a cookie in his hand. and that's it.]} \\
\midrule
Output & 
\texttt{The description is identified as \textbf{[B.Control]}.} \\
\bottomrule
\end{tabular}
\caption{LoRA setup for classification tasks.}
\label{tab:lora_classification}
\end{table*}

\begin{table*}[!ht]
\small
\centering
\begin{tabular}{lp{0.75\linewidth}}
\toprule
\textbf{Prompt Component} & \textbf{Content Example} \\
\midrule
System Prompt&
\texttt{You are a volunteer for a picture description task. \newline Below is the given comprehensive description of the task:\newline "The picture depicts a chaotic kitchen scene. On the left, a young girl is pointing upwards towards a boy who is standing on a stool and reaching for a cookie jar labeled 'COOKIE JAR' in an open cupboard. The boy is losing his balance and appears to be falling. In the center, a woman is calmly washing dishes at a sink that is overflowing with water, which is spilling onto the floor. To the right, there is a counter with plates and a bowl. Outside the window above the sink, there are trees and possibly a house in the distance." \newline You are required to describe the picture based on your specified mental condition.} \\
\midrule
Input & 
\texttt{\textbf{AD Instruction} \newline ['You are in dementia, and you are required to describe the picture.', 'You are an Alzheimer\'s patient, and you are required to describe the picture.', 'You are a patient with Alzheimer\'s disease, and you are required to describe the picture.', 'You are a patient with dementia, and you are required to describe the picture.', 'You are a patient with Alzheimer\'s, and you are required to describe the picture.', 'You are a patient with dementia, and you are required to describe the picture.', 'You are a patient with Alzheimer\'s disease, and you are required to describe the picture.', 'You are a patient with Alzheimer\'s, and you are required to describe the picture.', 'You are a patient with dementia, and you are required to describe the picture.', 'You are a patient with Alzheimer\'s disease, and you are required to describe the picture.'] \newline \textbf{Control Instruction} \newline \textit{['You are mentally healthy, and you are required to describe the picture.', 'You are a normal person, and you are required to describe the picture.', 'You are a person with normal mental health, and you are required to describe the picture.']} \newline You can start your description now.} \\
\midrule
Output & 
\texttt{\#Certain Transcripts in Training Dataset\#} \\
\bottomrule
\end{tabular}
\caption{LoRA setup for generation tasks.}
\label{tab:lora_generation}
\end{table*}

\begin{table*}[t]
\centering
\small
\setlength{\tabcolsep}{1pt}
\begin{tabular}{l|c|cccc|cccc|cccc}
\toprule
\multirow{2}{*}{\textbf{Methods}} & \multirow{2}{*}{\textbf{N}} &
\multicolumn{4}{c|}{\textbf{Test}} & \multicolumn{4}{c|}{\textbf{Lu}} & \multicolumn{4}{c}{\textbf{Pitt}} \\
 &  & Pre & Rec & F1 & Acc & Pre & Rec & F1 & Acc & Pre & Rec & F1 & Acc \\
\midrule
\multirow{5}{*}{\textbf{ICL$_{Van}$}} 
 & 0 & 60.73$_{9.36}$ & 45.00$_{7.17}$ & 51.50$_{7.34}$ & 57.71$_{6.46}$ 
     & 67.45$_{7.08}$ & 60.00$_{8.08}$ & 63.41$_{7.24}$ & 55.71$_{8.05}$ 
     & 63.28$_{2.34}$ & 49.02$_{1.97}$ & 55.23$_{2.00}$ & 55.72$_{1.96}$ \\
 & 1 & 66.07$_{5.57}$ & 73.33$_{5.65}$ & 69.33$_{4.22}$ & 67.50$_{5.12}$ 
     & 70.46$_{2.82}$ & 83.70$_{5.02}$ & 76.45$_{3.03}$ & 66.90$_{4.05}$ 
     & 67.03$_{1.03}$ & 67.35$_{1.58}$ & 67.18$_{1.11}$ & 63.33$_{1.12}$ \\
 & 2 & 65.15$_{3.34}$ & 72.08$_{6.19}$ & 68.29$_{3.71}$ & 66.67$_{3.36}$ 
     & 73.28$_{3.55}$ & 86.67$_{4.74}$ & 79.31$_{2.86}$ & 70.95$_{4.10}$ 
     & 68.87$_{0.91}$ & 66.67$_{2.13}$ & 67.74$_{1.35}$ & 64.63$_{1.15}$ \\
 & 3 & 66.01$_{2.86}$ & 76.25$_{3.75}$ & 70.66$_{1.83}$ & 68.33$_{2.43}$ 
     & 72.82$_{4.46}$ & 88.52$_{5.09}$ & 79.83$_{4.10}$ & 71.19$_{6.25}$ 
     & 69.21$_{1.46}$ & 68.43$_{1.61}$ & 68.81$_{1.40}$ & 65.43$_{1.55}$ \\
 & 4 & 67.31$_{3.48}$ & 77.92$_{6.47}$ & 72.13$_{4.15}$ & 70.00$_{4.08}$ 
     & 72.80$_{3.74}$ & 87.78$_{4.70}$ & 79.53$_{3.58}$ & 70.95$_{5.08}$ 
     & 69.54$_{1.14}$ & 70.98$_{1.56}$ & 70.24$_{1.12}$ & 66.48$_{1.20}$ \\
\midrule
\multirow{4}{*}{\textbf{ICL$_{Sem}$}} 
 & 1 & 64.23$_{3.46}$ & 74.58$_{5.73}$ & 68.93$_{3.88}$ & 66.46$_{4.00}$ 
     & 72.99$_{2.50}$ & 83.70$_{5.02}$ & 77.89$_{2.69}$ & 69.52$_{3.33}$ 
     & 68.49$_{1.16}$ & 66.73$_{1.73}$ & 67.59$_{1.21}$ & 64.34$_{1.21}$ \\
 & 2 & 65.36$_{5.38}$ & 72.92$_{7.74}$ & 68.70$_{5.14}$ & 66.88$_{5.39}$ 
     & 71.84$_{2.77}$ & 85.56$_{4.81}$ & 78.01$_{2.65}$ & 69.05$_{3.53}$ 
     & 69.93$_{1.73}$ & 66.57$_{2.68}$ & 68.18$_{1.83}$ & 65.39$_{1.79}$ \\
 & 3 & 68.37$_{5.90}$ & 80.42$_{6.19}$ & 73.71$_{4.73}$ & 71.25$_{5.65}$ 
     & 69.92$_{1.98}$ & 90.37$_{4.12}$ & 78.81$_{2.54}$ & 68.81$_{3.44}$ 
     & 70.98$_{1.06}$ & 72.45$_{1.36}$ & 71.70$_{0.83}$ & 68.12$_{0.93}$ \\
 & 4 & 67.21$_{4.82}$ & 83.33$_{6.45}$ & 74.20$_{3.69}$ & 71.04$_{4.32}$ 
     & 72.54$_{2.79}$ & 94.44$_{2.48}$ & 82.02$_{2.11}$ & 73.33$_{3.50}$ 
     & 71.25$_{0.99}$ & 76.57$_{2.33}$ & 73.79$_{1.21}$ & 69.71$_{1.13}$ \\
\midrule
\multirow{5}{*}{\textbf{ICL$_{Ens}$}} 
 & 0 & 62.57$_{7.56}$ & 49.58$_{10.28}$ & 54.71$_{7.20}$ & 59.58$_{5.76}$ 
     & 68.03$_{4.26}$ & 60.00$_{7.73}$ & 63.59$_{5.59}$ & 56.19$_{5.65}$ 
     & 64.73$_{2.34}$ & 49.84$_{2.95}$ & 56.30$_{2.67}$ & 56.92$_{2.21}$ \\
 & 1 & 61.63$_{5.76}$ & 67.92$_{6.19}$ & 64.43$_{4.62}$ & 62.50$_{5.19}$ 
     & 71.33$_{3.34}$ & 85.56$_{6.30}$ & 77.71$_{3.99}$ & 68.57$_{5.19}$ 
     & 66.93$_{1.41}$ & 67.91$_{2.49}$ & 67.41$_{1.89}$ & 63.42$_{1.83}$ \\
 & 2 & 68.70$_{3.72}$ & 74.58$_{5.42}$ & 71.42$_{3.71}$ & 70.21$_{3.73}$ 
     & 72.02$_{3.06}$ & 87.41$_{4.74}$ & 78.91$_{3.12}$ & 70.00$_{4.29}$ 
     & 70.04$_{1.57}$ & 68.20$_{1.39}$ & 69.11$_{1.40}$ & 66.01$_{1.59}$ \\
 & 3 & 67.47$_{4.84}$ & 73.33$_{6.77}$ & 70.20$_{5.24}$ & 68.96$_{5.31}$ 
     & 73.23$_{2.52}$ & 89.26$_{4.52}$ & 80.44$_{3.24}$ & 72.14$_{4.27}$ 
     & 70.10$_{1.06}$ & 68.82$_{1.57}$ & 69.44$_{0.93}$ & 66.25$_{0.95}$ \\
 & 4 & 66.75$_{5.05}$ & 86.67$_{4.08}$ & 75.32$_{4.08}$ & 71.46$_{5.44}$ 
     & 72.16$_{3.73}$ & 89.63$_{4.32}$ & 79.88$_{3.28}$ & 70.95$_{4.97}$ 
     & 70.93$_{1.16}$ & 70.42$_{1.69}$ & 70.66$_{1.07}$ & 67.41$_{1.10}$ \\
\midrule
\multirow{4}{*}{\textbf{TV$_{Van}^{Add}$}} 
 & 1 & 55.29$_{6.44}$ & 47.92$_{7.28}$ & 51.23$_{6.62}$ & 54.58$_{5.88}$ 
     & 67.78$_{6.05}$ & 55.93$_{8.83}$ & 61.10$_{7.41}$ & 54.76$_{6.82}$ 
     & 62.01$_{1.64}$ & 50.00$_{3.01}$ & 55.33$_{2.24}$ & 55.06$_{1.62}$ \\
 & 2 & 59.64$_{6.13}$ & 50.42$_{5.73}$ & 54.60$_{5.77}$ & 58.13$_{5.22}$ 
     & 68.22$_{4.91}$ & 57.41$_{8.49}$ & 62.12$_{6.34}$ & 55.48$_{6.03}$ 
     & 60.93$_{2.32}$ & 48.30$_{2.66}$ & 53.86$_{2.35}$ & 53.92$_{2.11}$ \\
 & 3 & 57.79$_{8.36}$ & 46.25$_{6.57}$ & 51.04$_{6.17}$ & 55.62$_{6.79}$ 
     & 67.37$_{4.76}$ & 51.85$_{8.28}$ & 58.42$_{6.72}$ & 53.10$_{5.84}$ 
     & 60.95$_{1.22}$ & 50.36$_{3.41}$ & 55.10$_{2.33}$ & 54.35$_{1.37}$ \\
 & 4 & 57.89$_{6.25}$ & 51.67$_{5.65}$ & 54.47$_{5.21}$ & 56.88$_{4.85}$ 
     & 68.39$_{7.75}$ & 57.04$_{8.31}$ & 62.06$_{7.81}$ & 55.48$_{8.11}$ 
     & 60.56$_{2.35}$ & 47.88$_{2.72}$ & 53.45$_{2.44}$ & 53.57$_{2.09}$ \\
\midrule
\multirow{4}{*}{\textbf{TV$_{Sem}^{Add}$}} 
 & 1 & 58.82$_{7.06}$ & 48.33$_{2.53}$ & 52.28$_{8.33}$ & 57.08$_{5.12}$ 
     & 72.66$_{5.87}$ & 58.89$_{8.52}$ & 64.79$_{6.62}$ & 59.29$_{6.43}$ 
     & 61.26$_{1.78}$ & 49.44$_{1.83}$ & 54.71$_{1.72}$ & 54.39$_{1.62}$ \\
 & 2 & 64.11$_{7.93}$ & 50.42$_{10.45}$ & 56.02$_{8.84}$ & 61.04$_{6.72}$ 
     & 69.37$_{4.21}$ & 60.74$_{6.67}$ & 64.54$_{4.53}$ & 57.38$_{4.32}$ 
     & 60.57$_{2.77}$ & 49.25$_{2.37}$ & 54.31$_{2.38}$ & 53.83$_{2.44}$ \\
 & 3 & 58.64$_{5.98}$ & 46.25$_{9.58}$ & 51.44$_{7.95}$ & 57.08$_{4.86}$ 
     & 67.11$_{3.80}$ & 57.04$_{6.67}$ & 61.55$_{5.31}$ & 54.52$_{4.70}$ 
     & 60.50$_{2.52}$ & 49.18$_{3.07}$ & 54.22$_{2.65}$ & 53.77$_{2.31}$ \\
 & 4 & 61.47$_{8.01}$ & 52.08$_{8.39}$ & 56.31$_{7.99}$ & 59.79$_{6.85}$ 
     & 68.96$_{4.64}$ & 56.30$_{6.58}$ & 61.91$_{5.65}$ & 55.71$_{5.65}$ 
     & 60.10$_{2.11}$ & 46.80$_{2.08}$ & 52.60$_{1.80}$ & 53.01$_{1.71}$ \\
\midrule
\multirow{4}{*}{\textbf{TV$_{Van}^{Rep}$}} 
 & 1 & 60.34$_{2.72}$ & 75.83$_{5.83}$ & 67.11$_{3.12}$ & 62.92$_{3.46}$ 
     & 68.80$_{3.24}$ & 83.33$_{6.47}$ & 75.31$_{4.16}$ & 65.00$_{5.43}$ 
     & 61.89$_{1.21}$ & 72.88$_{1.83}$ & 66.92$_{1.21}$ & 59.85$_{1.43}$ \\
 & 2 & 62.72$_{4.08}$ & 75.42$_{6.02}$ & 68.40$_{4.27}$ & 65.21$_{4.66}$ 
     & 70.32$_{2.4}$ & 82.96$_{3.39}$ & 76.06$_{1.71}$ & 66.43$_{2.49}$ 
     & 61.21$_{0.96}$ & 72.35$_{1.74}$ & 66.31$_{1.15}$ & 59.03$_{1.27}$ \\
 & 3 & 59.25$_{3.55}$ & 77.08$_{7.28}$ & 66.94$_{4.72}$ & 62.08$_{4.82}$ 
     & 70.46$_{3.35}$ & 84.44$_{5.19}$ & 76.75$_{3.43}$ & 67.14$_{4.86}$ 
     & 61.53$_{1.30}$ & 73.79$_{2.03}$ & 67.10$_{1.44}$ & 59.67$_{1.71}$ \\
 & 4 & 60.71$_{4.91}$ & 75.00$_{4.93}$ & 66.97$_{3.96}$ & 62.92$_{5.34}$ 
     & 67.64$_{3.25}$ & 82.96$_{6.46}$ & 74.46$_{4.28}$ & 63.57$_{5.33}$ 
     & 62.07$_{0.72}$ & 73.86$_{1.57}$ & 67.45$_{1.02}$ & 60.27$_{1.06}$ \\
\midrule
\multirow{4}{*}{\textbf{TV$_{Sem}^{Rep}$}} 
 & 1 & 59.85$_{5.53}$ & 71.68$_{7.64}$ & 65.00$_{4.94}$ & 61.46$_{5.76}$ 
     & 68.09$_{2.09}$ & 81.48$_{6.42}$ & 74.10$_{3.48}$ & 63.57$_{3.99}$ 
     & 61.55$_{0.75}$ & 73.53$_{1.00}$ & 67.00$_{0.72}$ & 59.64$_{0.91}$ \\
 & 2 & 57.04$_{2.16}$ & 74.17$_{4.49}$ & 64.46$_{2.86}$ & 59.17$_{2.83}$ 
     & 67.68$_{2.69}$ & 82.96$_{4.74}$ & 74.51$_{3.19}$ & 63.57$_{4.27}$ 
     & 62.31$_{1.40}$ & 73.99$_{2.58}$ & 67.63$_{1.63}$ & 60.55$_{1.82}$ \\
 & 3 & 61.95$_{4.12}$ & 80.00$_{8.90}$ & 69.68$_{5.40}$ & 65.42$_{5.45}$ 
     & 67.50$_{2.85}$ & 81.48$_{5.74}$ & 73.77$_{3.54}$ & 62.86$_{4.54}$ 
     & 61.79$_{1.25}$ & 73.17$_{1.64}$ & 66.99$_{1.23}$ & 59.82$_{1.55}$ \\
 & 4 & 57.47$_{4.10}$ & 72.92$_{7.03}$ & 64.15$_{4.65}$ & 59.38$_{4.95}$ 
     & 69.56$_{3.55}$ & 85.19$_{4.38}$ & 76.54$_{3.46}$ & 66.43$_{5.05}$ 
     & 62.14$_{0.90}$ & 74.22$_{2.17}$ & 67.64$_{1.39}$ & 60.44$_{1.38}$ \\
\midrule
\textbf{Ours} & 4 & \textbf{84.55$_{3.02}$} & \textbf{87.92$_{4.14}$} & \textbf{86.11$_{1.92}$} & \textbf{85.83$_{1.91}$} 
     & \textbf{81.08$_{2.22}$} & \textbf{91.85$_{1.56}$} & \textbf{86.12$_{1.73}$} & \textbf{80.95$_{2.51}$} 
     & \textbf{86.34$_{0.50}$} & \textbf{74.93$_{0.71}$} & \textbf{80.23$_{0.42}$} & \textbf{79.42$_{0.39}$} \\
\bottomrule
\end{tabular}
\caption{Main experimental results on the Test, Lu, and Pitt datasets. Each entry reports mean$_{\text{std}}$ over ten runs.}
\label{tab:main_results_full}
\end{table*}

\end{document}